\documentclass[10pt,twocolumn,letterpaper]{article}
\usepackage[pagenumbers]{cvpr}

\definecolor{cvprblue}{rgb}{0.21,0.49,0.74}
\usepackage[pagebackref,breaklinks,colorlinks,allcolors=cvprblue]{hyperref}

\usepackage{fontawesome}
\usepackage[utf8]{inputenc} %
\usepackage[T1]{fontenc}    %
\usepackage{booktabs}       %
\usepackage{amsfonts}       %
\usepackage{nicefrac}       %
\usepackage{microtype}      %
\usepackage[table]{xcolor}  %
\usepackage{amsmath}
\usepackage{graphicx}
\usepackage{tabularx}
\usepackage{multirow}
\usepackage{enumitem}
\usepackage{subcaption}
\usepackage{wrapfig}
\usepackage{algorithm,algpseudocode}
\usepackage{pifont}
\usepackage{bm}
\usepackage{makecell}
\usepackage{arydshln}
\usepackage{utfsym}
\usepackage{placeins}

\DeclareMathOperator*{\argmax}{arg\,max}

\definecolor{tabfirst}{rgb}{1, 0.7, 0.7} %
\definecolor{tabsecond}{rgb}{1, 0.85, 0.7} %
\definecolor{tabthird}{rgb}{1, 1, 0.7} %

\begin{document}
\title{Name That Part: 3D Part Segmentation and Naming}
\author{%
 Soumava Paul$^{\star}$  \quad Prakhar Kaushik$^{\star\ddag}$ \quad
 Ankit Vaidya \quad Anand Bhattad \quad Alan Yuille\\[0.4em]
\small $\star$ Equal Contribution \hspace{0.5em}$\ddag$ Project Lead
\\[0.8em]
{\normalsize Johns Hopkins University, Baltimore, MD, USA}
\\
{\small \{spaul27, pkaushi1, avaidya7, bhattad, ayuille1\}@jh.edu}
}

\twocolumn[{%
\renewcommand\twocolumn[1][]{#1}%
\maketitle
\vspace{-1.5em}
\centering
\url{https://name-that-part.github.io}
\vspace{1em}
  \begin{center}
  \includegraphics[width=\textwidth]{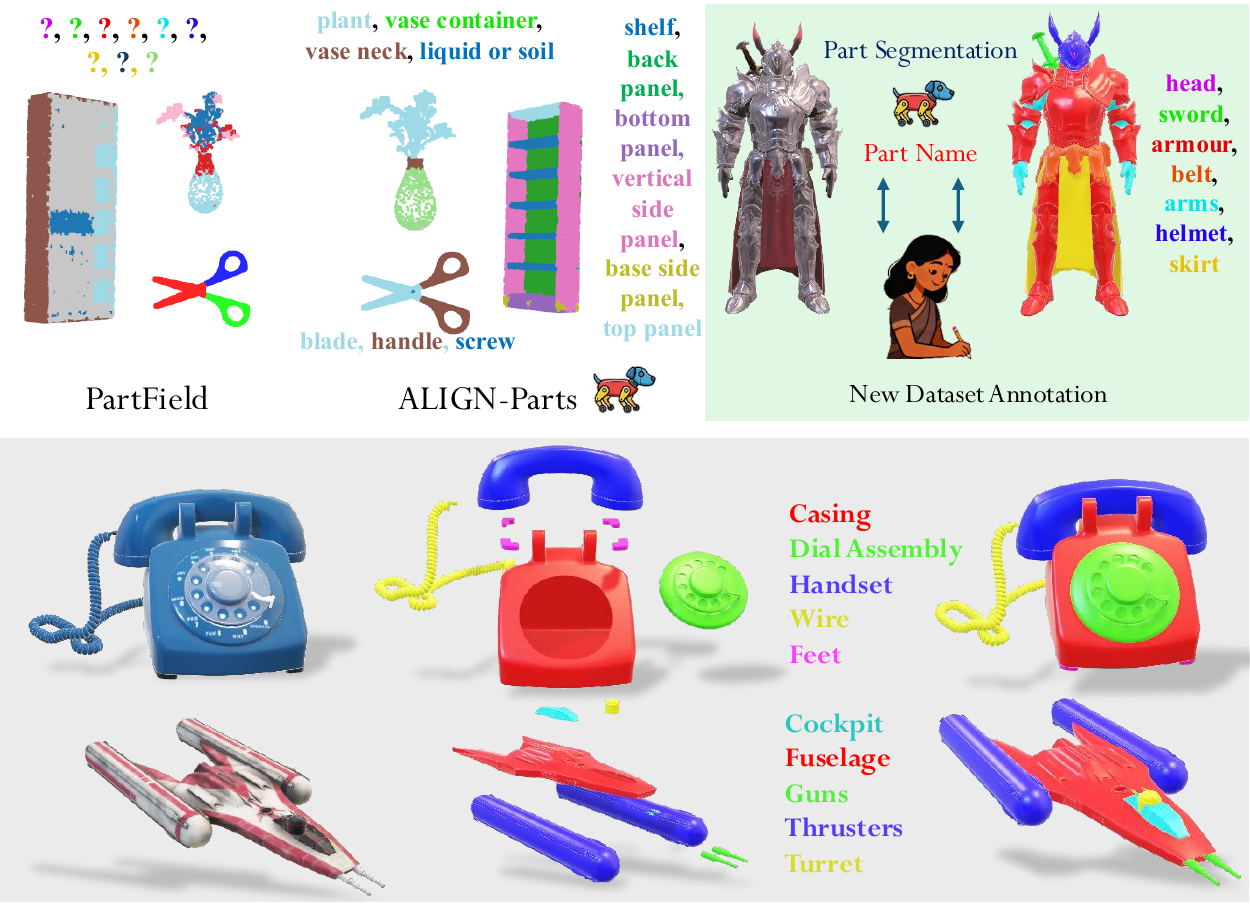}
  \captionof{figure}{\textbf{ALIGN-Parts} is the first large-scale method to be able to efficiently and semantically segment and name 3D parts of an object, unlike previous methods, which could only perform class-agnostic part segmentation. Our method is also feed-forward, and defines 3D parts according to human-oriented, object-specific affordance descriptions. (left) ALIGN-Parts segments and semantically names 3D parts, unlike PartField~\cite{liu2025partfield}, which only segments.  Our feed-forward method is faster at generating these segments along with names compared to PartField as we don't require clustering. (right) Simultaneous part segmentation and naming enable $5-8\times$ more efficient creation of 3D part datasets (bottom) with human-in-the-loop validation.} 
  \label{fig:teaser}
\end{center}
}]

\begin{abstract}
We address semantic 3D part segmentation: decomposing objects into parts with meaningful names. While datasets exist with part annotations, their definitions are inconsistent across datasets, limiting robust training. Previous methods produce unlabeled decompositions or retrieve single parts without complete shape annotations. We propose ALIGN-Parts, which formulates part naming as a direct set alignment task. 
Our method decomposes shapes into partlets - implicit 3D part representations - matched to part descriptions via bipartite assignment.  We combine geometric cues from 3D part fields, appearance cues from multi-view vision features, and semantic knowledge from language-model-generated affordance descriptions.
Text-alignment loss ensures partlets share embedding space with text, enabling a theoretically open-vocabulary matching setup, given sufficient data. Our efficient and novel, one-shot, 3D part segmentation and naming method finds applications in several downstream tasks, including serving as a scalable annotation engine. As our model supports zero-shot matching to arbitrary descriptions and confidence-calibrated predictions for known categories, with human verification, we create a unified ontology that aligns PartNet, 3DCoMPaT++, and Find3D, consisting of 1,794 unique 3D parts. We introduce two novel metrics appropriate for the named 3D part segmentation task. We also show examples from our newly created TexParts dataset.

\end{abstract}

\begin{table*}[!htbp]
\centering
\small
\begin{tabular}{lcccccc}
\toprule
Method & Complete & Named & Open & Permutation & Part-level & Feed\\
       & Decomposition  & Parts & Vocabulary & Invariant & Features & Forward\\
\midrule
PartField & \textcolor{green}{✓} & \textcolor{red}{✗} & \textcolor{red}{✗} & - & \textcolor{red}{✗} & \textcolor{red}{✗} \\
SAMPart3D & \textcolor{red}{✗} & \textcolor{red}{✗} & \textcolor{red}{✗} & \textcolor{red}{✗} & \textcolor{red}{✗} & \textcolor{red}{✗} \\
Find3D & \textcolor{red}{✗} & \textcolor{green}{✓} & \textcolor{green}{✓} & \textcolor{red}{✗} & \textcolor{red}{✗} & \textcolor{red}{✗} \\
\textbf{Ours} & \textcolor{green}{✓} & \textcolor{green}{✓} & \textcolor{green}{✓} & \textcolor{green}{✓} & \textcolor{green}{✓} & \textcolor{green}{✓} \\
\bottomrule
\end{tabular}
\caption{Comparison with 3D Part segmentation or retrieval methods.}
\label{tab:comparison}
\end{table*}
\section{Introduction}
\label{sec:intro}
Many vision tasks require 3D parts, not just whole-object labels. Examples include robots grasping handles and creators editing assets. This requires solving two problems simultaneously: geometrically segmenting the parts and semantically naming them. While large datasets of 3D objects exist, only a few provide part annotations, and these annotations are often inconsistent across datasets~\cite{Mo_2019_CVPR, slim20253dcompat++, ma2025find}. An algorithm that can provide accurate and consistent annotations of named parts on any 3D object would enable scalable training data and support human-in-the-loop annotation pipelines.

Existing methods address only one aspect of this problem. Part segmentation models can identify geometric boundaries but produce unnamed regions~\cite{liu2025partfield}; however, these arbitrary segmentations lack any semantic grounding (from a part-based human perspective). Language-grounded systems can retrieve a single part from a text query but fail to produce a complete set of names for all parts of an object~\cite{ma2025find}. Classical unsupervised part discovery lacks the semantic grounding to derive consistent definitions. This gap has created a bottleneck: the absence of large-scale, consistently-annotated 3D part data. Inducing consistent parts from unlabeled web assets requires coupling geometric features and semantic knowledge with human verification.

We propose ALIGN-Parts, which formulates 3D part naming as a direct set alignment problem. Rather than deciding per-point which text label to assign, we decompose the shape into a small set of shape-conditioned \textit{partlets}. Each partlet consists of a set of points (a segmentation mask) and a text description (embedding) corresponding to one part. These partlets aggregate information across all points belonging to a part: a single point on a chair seat contains limited information, but the set of all points on the seat specifies the part. We align this set of partlets to a set of candidate part descriptions via bipartite matching in a single forward pass. Each partlet inherits a name from its matched description. A ``null'' class allows the model to discard unused partlets, enabling the number of parts to adapt per shape while ensuring permutation consistency: each predicted part receives at most one name, and each name is used at most once.

Our key contribution is the application of set-level matching to 3D part fields. This enables lower computational complexity: we match a handful of partlets to descriptions instead of all points to all labels. It provides permutation consistency: each predicted part receives at most one name, and each name is used at most once. Because of this, ALIGN-Parts is significantly faster at generating these segments, along with their names.

To ensure partlets are both geometrically separable and semantically meaningful, we combine three signals - 
\begin{enumerate}
    \item Geometry features from a 3D part-field backbone~\cite{liu2025partfield} capture shape structure,
    \item Appearance features from multi-view image encoders~\cite{oquab2023dinov2} lifted onto 3D geometry provide texture cues, 
    \item Semantic knowledge comes from affordance-aware part descriptions that encode form-and-function relationships~\cite{connell1987generating}. For example, a chair seat becomes ``the horizontal surface where a person sits,'' linking its flat, horizontal geometry to its sitting affordance.
\end{enumerate}

For semantic text grounding, we generate affordance and category-based part descriptions using a large language model~\cite{comanici2025gemini} and embed them using MPNet sentence transformers~\cite{song2020mpnet}. By representing part descriptions as embeddings in a continuous space, our approach supports scenarios where the model can match partlets to any user-provided set of part descriptions without retraining. In order to evaluate the task of simultaneous 3D part segmentation and naming, we introduce two new metrics.

Our approach provides a tool to address the 3D part data bottleneck. We construct a unified part ontology using a hybrid LLM-and-human process that normalizes labels and verifies geometric consistency across PartNet, 3DCoMPaT++, and Find3D. We then apply ALIGN-Parts to bootstrap annotations from unlabeled TexVerse assets~\cite{zhang2025texverse}, creating Tex-Parts: a dataset with 8450 objects spanning 14k part categories. In this setup, ALIGN-Parts serves as a \textit{scalable annotation engine} that proposes named parts for human verification, converting raw meshes into training data with minimal effort and enabling the construction of large, consistently-annotated 3D datasets.

\noindent 
In summary, our contributions are: 
\begin{itemize}[leftmargin=*,nosep]
  \item \textbf{Direct 3D parts alignment for open-world part naming.}
  We introduce \textit{partlets} - shape-conditioned part proposals with text embeddings - and match them to candidate descriptions via bipartite assignment, enabling efficient labeling of 3D part segmentation and naming. 
  
  \item \textbf{Geometry-appearance-semantic fusion.}
  We combine geometric structure, appearance features, and affordance-aware LLM-generated descriptions to produce semantically grounded, visually coherent partlets.

  \item \textbf{Metrics for evaluating 3D part semantic segmentation} We introduce 2 metrics suitable for our 3D part segmentation and naming task, which evaluate ALIGN-Parts and related baselines for both part segmentation accuracy and semantic correctness of the predicted part label.
  
  \item \textbf{Unified ontology and scalable annotation engine.}\\
  We harmonize part taxonomies across PartNet, 3DCoMPaT++, and Find3D datasets, and demonstrate a human-in-the-loop pipeline that bootstraps \textbf{TexParts}, a verified benchmark of 8450 objects spanning 14k categories derived from Texverse~\cite{zhang2025texverse}. ALIGN-Parts converts raw meshes into training data with $5$-$8\times$ less human effort than manual annotation.
\end{itemize}

\section{Related Work}
\label{sec:related}

\paragraph{3D Part Segmentation.}
Traditional methods operate in a purely geometric regime~\cite{10.1007/s00371-007-0197-5, 10.1145/1409060.1409098}. Early methods rely on handcrafted features and geometric consistency to partition shapes into meaningful regions, without assigning semantic names. Recent advances, such as PartField~\cite{liu2025partfield}, the current state-of-the-art, learns dense per-point feature fields but produce \textit{unlabeled} regions. 

More recent works lift 2D foundation models into 3D: SAM-based approaches~\cite{ma2025p3sam, yang2024sampart3d, tang2024segment} adapt Segment-Anything via multi-view projection but require prompts and lack semantic names. PartSTAD~\cite{kim2024partstad} integrates GLIP and SAM with 3D-aware objectives, while Diff3F~\cite{dutt2024diffusion} exploits diffusion features for segmentation. However, these methods operate as multi-stage pipelines and do not produce complete, non-overlapping part decompositions with coherent semantic 3D part names.

Among prior works, \citet{Kalogerakis:2010:labelMeshes, Kalogerakis2017} are the closest to our problem setting, as they explicitly formulate joint 3D part segmentation and semantic labeling. 
\cite{Kalogerakis:2010:labelMeshes} assigns fixed semantic labels to mesh faces using a Conditional Random Field with handcrafted geometric features, while the \cite{Kalogerakis2017} extension introduces learned features via projective convolutions. Despite their importance, these approaches rely on closed, category-specific label sets and lack the ability to scale to open-vocabulary naming or instance-consistent part identities. Moreover, labeling is framed as classification over predefined semantics rather than as a language-grounded naming problem.

As a result, most subsequent work has focused on amodal or class-based part segmentation, avoiding the challenges of assigning coherent and permutation-consistent semantic names to discovered parts.
Other works related to 3D part analysis, but not directly addressing part naming, include shape correspondence, retrieval, and structural understanding methods~\cite{ jones2020shapeAssembly, hanocka2019meshcnn, abstractionTulsiani17, Stelzner2021Decomposing3S, takmaz2023openmask3d}.

\paragraph{Language-Grounded 3D Understanding.}
PartSLIP~\cite{liu2023partslip}, PartSLIP++~\cite{zhou2023partslip++}, and PartDistill~\cite{umam2024partdistill} use image-language models for part segmentation but require per-category fine-tuning with predefined vocabularies. Find3D~\cite{ma2025find} and PartGlot~\cite{koo2022partglot} are most related: Find3D trains a point transformer in text embedding space for text-to-part retrieval. However, Find3D operates query-by-query (given "wing, head," it returns masks) rather than producing complete decompositions, and works on individual point features rather than part-level aggregations. PartGlot derives part-level
segmentation masks from spatial attention as a byproduct of learning to play the language reference game. However, it is formulated as a choice game for only a single class of object (chair), and a predefined small number of parts.

\paragraph{ALIGN-Parts} ALIGN-Parts differs fundamentally from prior work by formulating part segmentation and naming as a \textit{set alignment problem}: Partlets aggregate point features into part-level representations matched to text via optimal transport. This formulation enables: 
(1) complete, non-overlapping part decompositions in a single forward pass, 
(2) permutation-consistent semantic naming, 
(3) dynamic part cardinality without predefined part-count, and 
(4) zero-shot generalization across categories and datasets (\cref{tab:comparison}). 
Unlike prior multi-stage or query-based methods, ALIGN-Parts jointly learns segmentation and semantic alignment in an end-to-end manner. The number of parts emerges automatically from activated Partlets, eliminating the need for explicit part-count prediction or part-name inputs at inference time.

\paragraph{Part-Based Generation and Datasets.}
Part-based generative models~\cite{chen2025partgen, chen2025autopartgen, tang2024partpacker, lin2025partcrafter, yang2025holopart} discover latent part structures through synthesis, but do not address open-vocabulary semantic naming. Existing datasets such as PartNet~\cite{Mo_2019_CVPR}, 3DCoMPaT++~\cite{li20223d_compat, slim20253dcompat++}, and GAPartNet~\cite{geng2023gapartnet} provide part annotations, but rely on inconsistent taxonomies and category-specific semantics. ALIGN-Parts constructs a unified ontology across these datasets, enabling cross-dataset evaluation under a shared semantic framework.

\begin{figure*}[!htbp]
    \centering
    \includegraphics[width=\linewidth]{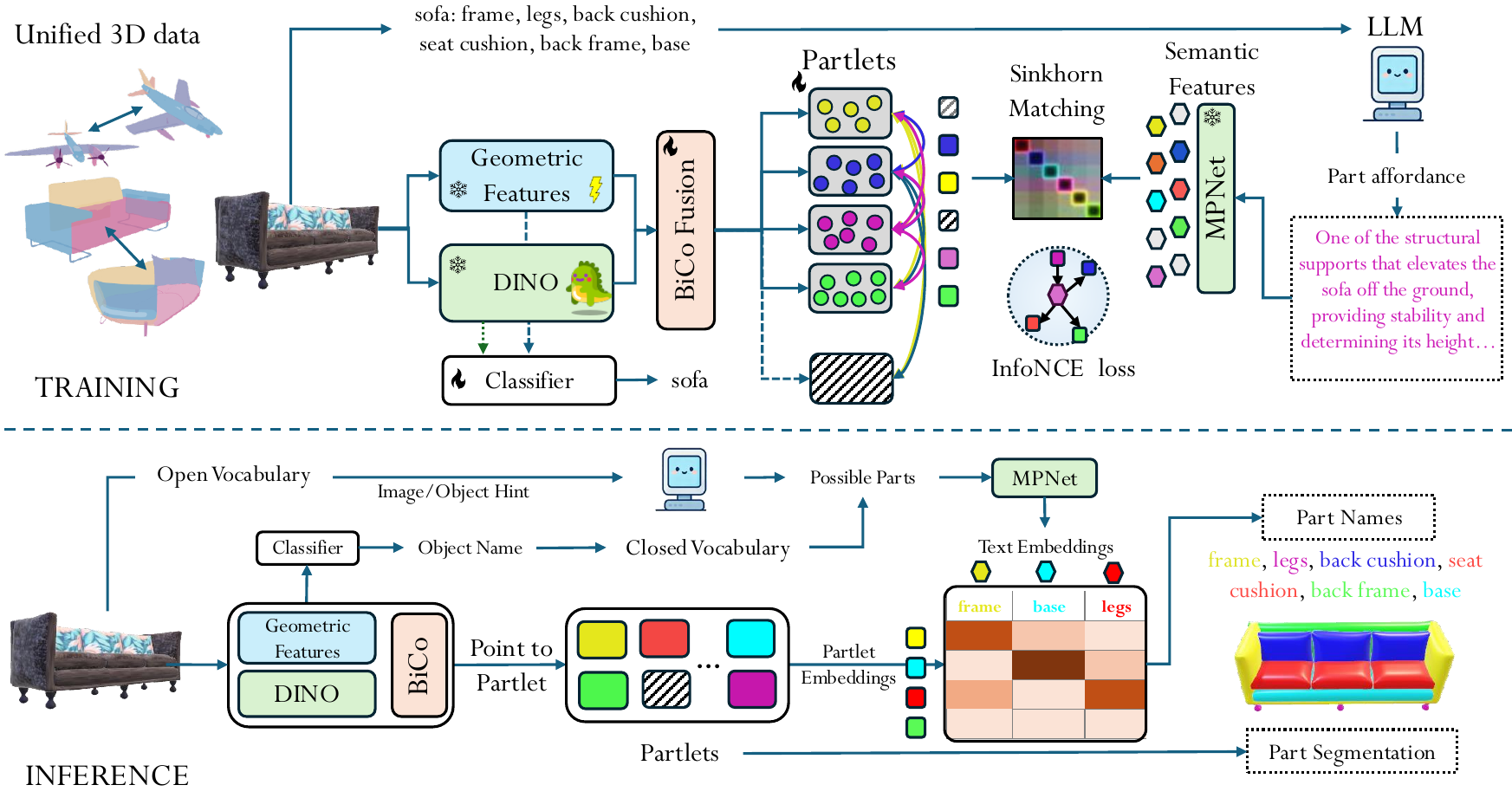} 

\caption{\textbf{ALIGN-Parts.} Overview of the ALIGN-Parts framework for language-grounded 3D part segmentation and naming. \textit{Top: training.} Given a 3D from our semantically unified 3D parts data, geometry features are extracted with PartField and appearance features with DINOv2 from multi-view renderings; these are fused by the BiCo Fusion module using efficient bi-directional cross-attention on local \(k{=}16\) nearest-neighbor graphs in 3D space, reducing complexity from \(\mathcal{O}(N^2)\) to \(\mathcal{O}(Nk)\) and yielding enriched point features. A decoder then learns \(K\) part-level ``Partlet'' representations that aggregate the fused features, with segmentation supervision provided at the Partlet level. To semantically ground Partlets, an LLM generates affordance-aware descriptions for each possible part (e.g., ``structural supports that elevate the sofa'' for sofa legs), which are embedded by a pretrained MPNet encoder; Sinkhorn matching establishes a bipartite assignment between Partlet and text embeddings, and an InfoNCE loss further aligns them in a shared representation space while a classifier predicts the object category. \textit{Bottom: inference.} At test time, ALIGN-Parts operates in both closed-vocabulary (object categories similar to those seen in training) and open-vocabulary (novel object categories) settings: in the closed-vocabulary case, the trained 3D classifier predicts the object class and retrieves its candidate part list, whereas in the open-vocabulary case an LLM proposes an overcomplete set of plausible parts for the queried object. Given these candidate part texts, their MPNet embeddings are bipartite-matched to the predicted Partlets, which jointly produce 3D part segmentation masks and corresponding part names.}
\label{fig:method_full}
\end{figure*}

\section{Method}
\label{sec:method}
\paragraph{Overview.} We propose ALIGN-Parts, a framework that treats automatic semantic 3D part segmentation as a direct set alignment problem, analogous to DETR~\cite{carion2020end} in 2D detection. The pipeline consists of three components:
\begin{enumerate}
    \item A dense feature fusion module combining geometry and appearance via localized bi-directional 3D-aware cross-attention,
    \item A Partlets module that learns K adaptive part-level representations by aggregating fused point features, and
    \item A semantic grounding module that aligns Partlets to part names via text descriptions of part affordances.
\end{enumerate}
At inference time, given a 3D shape and generated candidate part descriptions, our method produces named part segments without specifying part counts, names, or point prompts. Training uses contrastive alignment and differentiable optimal transport matching between predicted Partlets and ground-truth parts, enabling end-to-end learning of both segmentation and semantics.
\paragraph{Problem Formulation.}
Given a 3D point cloud $\mathcal{P} = \{\mathbf{p}_i\}_{i=1}^N$, each point has geometry-first features $\mathbf{f}_i^g \in \mathbb{R}^{d_g}$, appearance-first features $\mathbf{f}_i^a \in \mathbb{R}^{d_a}$, and coordinates $\mathbf{x}_i \in \mathbb{R}^3$. During training, we have ground-truth part masks $\mathbf{M}^{\text{gt}} \in \{0,1\}^{A \times N}$ and corresponding text embeddings $\{\mathbf{t}_a\}_{a=1}^A$ where $\mathbf{t}_a \in \mathbb{R}^{d_t}$ is an MPNet embedding of a language-model-generated semantic part description.

Our model learns $K=32$ instance-specific Partlet embeddings $\{\mathbf{s}_k \in \mathbb{R}^{d_t}\}_{k=1}^K$ that reside in the same dimensional space as semantic text embeddings (we use $d_t=768$). Each Partlet predicts three outputs: (i) a soft part mask $\mathbf{m}_k \in [0,1]^N$ over points, (ii) a partness score $\text{part}_k \in \mathbb{R}$ indicating whether it represents an actual part, and (iii) its embedding $\mathbf{s}_k$ which serves as the prototype for semantic alignment. We select $K=32$ to accommodate variable part configurations (We find that most 3D shapes have $\leq 28$ parts) while allowing future extensibility. Point part labels are obtained by matching Partlets to ground-truth or generated parts' descriptions via optimal transport.

\subsection{Architecture}
The input to our model is raw geometric and appearance features. We use PartField and multiview DiNO features; however, they can be replaced with any viable alternatives. Ideally, we want to train these input feature modules in an end-to-end manner with the rest of our model to extract optimal performance from the model. However, we are constrained due to compute requirements and lack of sufficient finetuning details~\cite{liu2025partfield}. Better performance than shown in this work is likely possible with better and trainable input modules.

\paragraph{Feature Fusion.}
The raw geometric features $\mathbf{f}_i^g$ and appearance features $\mathbf{f}_i^a$ capture complementary information: geometry encodes shape characteristics, while appearance provides texture and visual cues. We fuse these modalities through bi-directional cross-attention that we call BiCo Fusion operating on local $k{=}16$ nearest neighbor graphs in 3D coordinate space, reducing complexity to $\mathcal{O}(Nk)$.

For each point $i$ along with its nearest neighbors, geometric features attend to the appearance features of neighbors, producing cross-modal features that capture appearance information. Symmetrically, appearance features attend to geometric features of neighbors, capturing geometric information. We incorporate 3D spatial structure through Fourier-encoded relative positional biases. Learned sigmoid gates control how much of this cross-modal information to incorporate into each original feature, based on both the original feature and the attended information. After gated addition and layer normalization, we concatenate both modalities and project through a two-layer MLP to produce fused features $\mathbf{h}_i \in \mathbb{R}^{256}$ for each point. See \cref{sec_apx:featfusion} for details.

\paragraph{Points $\rightarrow$ Partlets: Learning Part-Level Representations.}
We learn $K$ Partlet embeddings that aggregate point-level information into part-level representations. A Partlet is defined by three components: (i) a soft segmentation mask $\mathbf{m} \in [0,1]^N$ representing membership scores, (ii) a Partlet embedding $\mathbf{s} \in \mathbb{R}^{d_t}$ in the learned semantic space, and (iii) a text embedding $\mathbf{z} \in \mathbb{R}^{d_t}$ representing the part description.

Formally, we learn a parameterized function $f_\theta: \mathbb{R}^{d_h \times N} \rightarrow (\mathbb{R}^N \times \mathbb{R}^{d_t})^K$ that maps fused point features $\mathbf{H} = \{\mathbf{h}_i\}_{i=1}^N$ to K Partlets:
\begin{equation}
(\mathbf{m}_k, \mathbf{s}_k) = f_\theta(\mathbf{H}), \quad k=1,\ldots,K
\end{equation}
The parameters $\theta$ include all weight matrices and biases in the refinement network described below. The motivation for Partlets is: individual point features cannot reliably map to semantic part descriptions (e.g., a single point on a chair seat lacks context to predict "seat"), but aggregating features across all points in a part enables robust semantic grounding.

We initialize K learnable Partlet embeddings $\{\mathbf{s}_k^{(0)} \in \mathbb{R}^{d_t}\}_{k=1}^K$ sampled from $\mathcal{N}(\mathbf{0}, \mathbf{I})$, shared across all shapes but adapted per instance through L refinement layers. At each layer $\ell$, each Partlet undergoes three operations:

\noindent\textbf{Partlet-to-Partlet Interaction:} Partlets interact to model part co-occurrence (e.g., chairs have seats and backs):
\begin{equation}
\mathbf{s}_k^{(\ell,1)} = \mathbf{s}_k^{(\ell-1)} + \text{SelfAttn}(\mathbf{s}_k^{(\ell-1)}, \{\mathbf{s}_{k'}\}_{k'=1}^K)
\end{equation}

\noindent\textbf{Point-to-Partlet Aggregation:} Partlets gather shape-specific evidence from BiCo-fused point features:
\begin{equation}
\mathbf{s}_k^{(\ell,2)} = \mathbf{s}_k^{(\ell,1)} + \text{CrossAttn}(\mathbf{s}_k^{(\ell,1)}, \{\mathbf{h}_i\}_{i=1}^N)
\end{equation} 

\noindent\textbf{Non-linear Transformation:} Two-layer MLP with GELU:
\begin{equation}
\mathbf{s}_k^{(\ell)} = \mathbf{s}_k^{(\ell,2)} + \text{MLP}(\mathbf{s}_k^{(\ell,2)})
\end{equation}

 After L layers, we obtain refined Partlet embeddings $\mathbf{s}_k = \mathbf{s}_k^{(L)}$.

\paragraph{Mask Prediction.}
Each Partlet predicts point membership via scaled dot-product:
\begin{equation}
m_{ki} = \frac{(\mathbf{W}_q \mathbf{s}_k)^T (\mathbf{W}_k \mathbf{h}_i)}{\sqrt{d_t}}
\end{equation}
During training, we apply sigmoid activation; at inference, softmax across Partlets yields soft assignments.

\paragraph{Partness Prediction.}
Each Partlet predicts whether it represents an actual part:
\begin{equation}
\text{part}_k = \mathbf{w}_{\text{part}}^T \mathbf{s}_k + b_{\text{part}}
\end{equation}
Higher values indicate active Partlets; lower values signify ``no-part,'' enabling dynamic part count adaptation per shape.

\paragraph{Semantic Alignment: Partlet $\rightarrow$ Part Names.}
Each Partlet's refined embedding $\mathbf{s}_k$ serves directly as its prototype $\mathbf{z}_k = \mathbf{s}_k \in \mathbb{R}^{d_t}$. After normalization ($\hat{\mathbf{z}}_k = \mathbf{z}_k / \|\mathbf{z}_k\|_2$), we compute cosine similarity with text embeddings $\hat{\mathbf{t}}_a$:

\begin{equation}
\text{sim}(k, a) = \hat{\mathbf{z}}_k \cdot \hat{\mathbf{t}}_a
\end{equation}

Importantly, Partlet and text embeddings share the same dimensional space ($\mathbb{R}^{d_t}$) without intermediate projections - this alignment is driven by the text alignment loss (\cref{sec:training}), enabling open-vocabulary matching at inference without retraining. However, part names are often ambiguous - \textit{handle} could refer to a door handle, a mug handle, or a wheelchair handle, each with distinct geometry and function. To disambiguate, we use affordance-based descriptions (e.g., ``the part of a door grasped to open it'' for a door handle, ``the horizontal surface of a chair where a person sits'' for a chair seat) generated by Gemini 2.5 Flash~\cite{comanici2025gemini}. These are embedded with a sentence transformer model, such as MPNet (all-mpnet-base-v2)~\cite{song2020mpnet}, for capturing long-form semantic affordances, particularly for similar part names. 

\subsection{Training: Partlet $\leftrightarrow$ Part Names}
\label{sec:training}

We establish correspondences between predicted Partlets and ground-truth parts via differentiable optimal transport. The cost matrix $\mathbf{C} \in \mathbb{R}^{K \times A}$ combines mask overlap and semantic similarity:
\begin{equation}
C_{ka} = \mathcal{L}_{\text{mask}}^{(k,a)} + (1 - \text{sim}(\hat{\mathbf{z}}_k, \hat{\mathbf{t}}_a))
\end{equation}
where $\mathcal{L}_{\text{mask}}^{(k,a)} = 1 - \text{Dice}(\sigma(\mathbf{m}_k), \mathbf{m}_a^{\text{gt}})$ (equal weighting $\alpha = \beta = 1.0$).

Sinkhorn-Knopp iterations produce a soft assignment matrix $\mathbf{P} \in [0,1]^{K \times A}$. Thresholding yields hard assignments $\pi: \{1,\ldots,K\} \rightarrow \{1,\ldots,A\} \cup \{\emptyset\}$, where $\pi(k) = a$ matches Partlet $k$ to part $a$ and $\pi(k) = \emptyset$ indicates no match.

\paragraph{Losses.} Let $\mathcal{M} = \{k : \pi(k) \neq \emptyset\}$ denote matched Partlets. Our training objective combines several losses:

\paragraph{Text Alignment Loss.}
This loss is essential for open-vocabulary grounding. Without it, Partlet embeddings remain geometrically meaningful but semantically ambiguous. We apply InfoNCE contrastive loss to align matched Partlets with their text embeddings:
\begin{equation}
\mathcal{L}_{\text{text}} = \frac{1}{|\mathcal{M}|} \sum_{k \in \mathcal{M}} -\log \frac{\exp(\hat{\mathbf{z}}_k \cdot \hat{\mathbf{t}}_{\pi(k)} / \tau)}{\sum_{a=1}^A \exp(\hat{\mathbf{z}}_k \cdot \hat{\mathbf{t}}_a / \tau)}
\end{equation}
with $\tau = 0.07$. Operating over Partlets rather than individual points makes this optimization tractable and stable.

\paragraph{Mask and Partness Losses.}
For matched Partlets, binary cross-entropy and Dice loss supervise masks:
\begin{equation}
\begin{aligned}
\mathcal{L}_{\text{mask}} = \frac{1}{|\mathcal{M}|} \sum_{k \in \mathcal{M}} [\text{BCE}(\mathbf{m}_k, \mathbf{m}_{\pi(k)}^{\text{gt}}) \\ +  (1 - \text{Dice}(\sigma(\mathbf{m}_k), \mathbf{m}_{\pi(k)}^{\text{gt}}))]
\end{aligned}
\end{equation}

Binary classification loss supervises partness, teaching Partlets to predict whether they are active (matched) or inactive (``no-part''):

\begin{equation}
\mathcal{L}_{\text{part}} = \frac{1}{K} \sum_{k=1}^K \text{BCE}(\text{part}_k, \mathbf{1}[\pi(k) \neq \emptyset])
\end{equation}

\paragraph{Auxiliary Regularizers.}
Coverage loss prevents over- / under-segmentation by penalizing mask size disparities:
\begin{equation}
\mathcal{L}_{\text{cov}} = \frac{1}{|\mathcal{M}|} \sum_{k \in \mathcal{M}} \left|\frac{\sum_i \sigma(m_{ki}) - \sum_i m_{\pi(k)i}^{\text{gt}}}{N}\right|
\end{equation}

Overlap loss enforces mutual exclusivity—each point should belong to at most one part:

\begin{equation}
\mathcal{L}_{\text{overlap}} = \frac{1}{N} \sum_{i=1}^N \left(\sum_{k=1}^K \sigma(m_{ki}) - 1\right)^2
\end{equation}

\paragraph{Global Alignment Loss.}
Symmetric InfoNCE aligns the global shape representation with class-level text embeddings, providing object-level semantic context:
\begin{equation}
\mathcal{L}_{\text{global}} = \frac{1}{2}\left[\mathcal{L}_{\text{CE}}(\mathbf{S}, \mathbf{I}) + \mathcal{L}_{\text{CE}}(\mathbf{S}^{\top}, \mathbf{I})\right]
\end{equation}
where $\mathbf{S} = \frac{1}{\tau} \hat{\mathbf{Z}}_{\text{global}} \hat{\mathbf{T}}_{\text{class}}^{\top}$.

\paragraph{Total Loss.}
The complete training objective as a function of model parameters $\theta$ is:
\begin{equation}
\begin{aligned}
\mathcal{L}_{\text{total}}(\theta) = \,
&\lambda_{\text{mask}}\mathcal{L}_{\text{mask}}(\theta, \pi) + \lambda_{\text{part}} \mathcal{L}_{\text{part}}(\theta, \pi) \\+ \lambda_{\text{text}} \mathcal{L}_{\text{text}}(\theta, \pi) 
&+ \lambda_{\text{cov}} \mathcal{L}_{\text{cov}}(\theta, \pi) + \lambda_{\text{ov}} \mathcal{L}_{\text{overlap}}(\theta, \pi) \\+ \lambda_{\text{global}} \mathcal{L}_{\text{global}}(\theta)
\end{aligned}
\end{equation}
where $\theta$ includes all learnable parameters (BiCo fusion weights, Partlet decoder layers, prediction heads), $\pi$ is the assignment from Partlets to ground-truth parts, computed via Sinkhorn matching given the current model predictions. All hyperparameters are in the supplement.

\subsection{Inference Modes and Use Cases}
\label{sec:inference}

Our model supports three primary inference modes for different deployment scenarios.\\

\noindent
\textbf{Mode 1: Closed-Vocabulary with Confidence Calibration.}
We use ALIGN-Parts to scalably annotate large datasets from known categories (e.g., labeling millions of airplane meshes). This is the most practical scenario, enabling efficient 3D part segmentation and labeling with minimal oversight. We utilize this capability in building the TexParts dataset (\cref{subsec:texparts}). For training categories $\mathcal{C}$, we predict the object category via global shape-text alignment:
\begin{equation}
\label{eq:inference_category_match}
c^* = \argmax_{c \in \mathcal{C}} \text{sim}(\mathbf{z}_{\text{global}}, \mathbf{t}_c)
\end{equation}
then filter Partlets by partness score ($\sigma(\text{part}_k){>}0.5$) to obtain active set $\mathcal{K}_{\text{active}}$. We construct cost matrix $\mathbf{C} \in \mathbb{R}^{|\mathcal{K}_{\text{active}}| \times |\mathcal{L}_{c^*}|}$:
\begin{equation}
C_{ka} = 1 - \text{sim}(\hat{\mathbf{z}}_k, \hat{\mathbf{t}}_a)
\end{equation}
At inference, we use the Jonker-Volgenant algorithm~\cite{7738348} for exact optimal assignment, which is more efficient than Sinkhorn since gradients are not required.

\paragraph{\textbf{Mahalanobis Score Estimation}}
The Mahalanobis confidence (\textbf{\cref{eq:inference_conf_maha}}) requires class-conditional statistics (mean and covariance) that are estimated from the training set.

After training, we perform a single forward pass over the entire training dataset. For every partlet $k$ that is successfully matched to a ground-truth part label $\ell$ (i.e., $\pi(k) = \ell$), we extract its prototype embedding $\mathbf{z}_k$.

We then compute the empirical mean $\bm{\mu}_\ell$ for each part label $\ell$ in our known training vocabulary $\mathcal{C}$:
\begin{equation}
\label{eq:supp_mean}
    \bm{\mu}_\ell = \mathbb{E}[\mathbf{z}_k | \pi(k) = \ell]
\end{equation}
For robustness, we compute a single, shared covariance matrix $\bm{\Sigma}$ by pooling the embeddings from all part classes:
\begin{equation}
\label{eq:supp_cov}
    \bm{\Sigma} = \text{Cov}(\{\mathbf{z}_k\}_{\forall k, \ell \text{ s.t. } \pi(k) = \ell})
\end{equation}
We apply regularization (e.g., adding a small value $\epsilon \mathbf{I}$ to the diagonal) before computing the inverse $\bm{\Sigma}^{-_1}$ to ensure numerical stability. These pre-computed $\bm{\mu}_\ell$ and $\bm{\Sigma}^{-_1}$ are stored and used at inference time to compute the following Mahalanobis-distance based confidence score:
\begin{equation}
\label{eq:inference_conf_maha}
\text{conf}_{\text{maha}}(k) = \exp(-(\mathbf{z}_k - \bm{\mu}_{a_k^*})^T \bm{\Sigma}_{a_k^*}^{-1} (\mathbf{z}_k - \bm{\mu}_{a_k^*}))
\end{equation}
where $\bm{\mu}_{a_k^*}$ and $\bm{\Sigma}_{a_k^*}$ are estimated from training embeddings. Predictions with $\text{conf}_{\text{maha}} >= 0.8$ are auto-accepted; lower-confidence predictions are flagged for human verification, dramatically reducing annotation cost.

\paragraph{\textbf{Fused Confidence Formulation}}
The final confidence score $\text{conf}(k)$ for a matched query $k$ is a fusion of the softmax confidence ($\text{conf}_{\text{soft}}$) and the Mahalanobis confidence ($\text{conf}_{\text{maha}}$). We combine them as follows:
\begin{equation}
\label{eq:supp_conf_fused}
    \text{conf}(k) = \alpha \cdot \text{conf}_{\text{soft}}(k) + (1-\alpha) \cdot \sigma(\beta \cdot (\text{conf}_{\text{maha}}(k) - 0.5))
\end{equation}
where $\sigma(\cdot)$ is the sigmoid function.

\begin{itemize}
    \item $\text{conf}_{\text{soft}}(k)$ is the temperature-calibrated softmax score (\textbf{\cref{eq:inference_conf_soft}}).
    \item $\text{conf}_{\text{maha}}(k)$ is the Mahalanobis confidence (\textbf{\cref{eq:inference_conf_maha}}).
    \item $\alpha$ and $\beta$ are hyperparameters that balance the two scores. We set $\alpha=0.5$ and $\beta=1.0$ based on calibration on a held-out validation set.
\end{itemize}

Annotations with $\text{conf}(k) < \tau_{\text{conf}}$ (where $\tau_{\text{conf}}=0.5$) are flagged as low-confidence and routed to a human annotator for manual review.\\

\noindent
\paragraph{Mode 2: Open-Vocabulary Grounding.}
For novel categories, users provide candidate part descriptions $\{\mathbf{t}_a\}_{a=1}^A$ (or generate via LLM from image/hint). After filtering inactive Partlets, each active Partlet matches to the best description:
\begin{equation}
\label{eq:inference_open_match}
a_k^* = \argmax_{a \in \{1,\ldots,A\}} \text{sim}(\hat{\mathbf{z}}_k, \hat{\mathbf{t}}_a)
\end{equation}
Confidence is computed via temperature-calibrated softmax:
\begin{equation}
\label{eq:inference_conf_soft}
\text{conf}_{\text{soft}}(k) = \max_a \frac{\exp(\hat{\mathbf{z}}_k \cdot \hat{\mathbf{t}}_a / \tau)}{\sum_{a'=1}^A \exp(\hat{\mathbf{z}}_k \cdot \hat{\mathbf{t}}_a' / \tau)}
\end{equation}
with $\tau = 0.07$. This is less calibrated than the Mahalanobis distance (Mode 1) due to the lack of training statistics for novel categories.\\

\paragraph{Mode 3: Text-Conditioned Part Retrieval.}
For comparison with Find3D~\cite{ma2025find}, we retrieve a single part for query $\mathbf{t}_q$:
\begin{equation}
\label{eq:inference_retrieval}
k^* = \argmax_{k \in \mathcal{K}_{\text{active}}} \text{sim}(\hat{\mathbf{z}}_k, \hat{\mathbf{t}}_q)
\end{equation}
returning mask $\mathbf{m}_{k^*}$. This mode is primarily for benchmark comparison.

\paragraph{\textbf{Point Label Assignment.}}
Points are assigned to the highest-scoring Partlet's label:
\begin{equation}
\text{label}_i = a_{k^*}^* \quad \text{where} \quad k^* = \argmax_{k \in \mathcal{K}_{\text{active}}} \sigma(m_{ki})
\end{equation}
Points with $\max_k \sigma(m_{ki}) < 0.5$ remain unlabeled.

\subsection{Implementation Details}

\paragraph{Training Setup.} ALIGN-Parts is trained on 3 NVIDIA A6000 GPUs for 2 days (batch size 16). \textbf{Due to academic compute constraints, we sample 10k points per shape (vs. 100k in PartField)} while maintaining strong performance.  Models are normalized to $[-1, 1]^3$ during training.

\paragraph{Architecture.} The model has 34M parameters total: 5.7M for feature Fusion, 26.8M for Partlets, and 1.5M for the global classifier. Feature interactions (Partlet-to-points and Partlet-to-Partlet) use 3 transformer~\cite{vaswani2017attention} blocks with multi-head cross-attention, LayerNorm, residual connections, and feedforward layers. The BiCo fusion employs sparse 16-NN attention with 3D relative positional bias, computed via a learned MLP over Fourier-encoded (F=6 frequencies) displacement vectors, which provides geometric context while maintaining $\mathcal{O}(Nk)$ complexity. We also note that for calculating the runtime, we do not include data preprocessing time, as it varies depending on parallelization and system capabilities.

We set the number of Partlets to 32, as this value provides a reasonable estimate for the typical number of semantic parts found in most objects in our unified dataset. This choice is further validated by analyzing the statistics of part counts across the full dataset, which confirms that 32 accommodates the majority of objects without excessive over-segmentation or loss of fine granularity.

\paragraph{Optimization.} Loss weights: $\lambda_{\text{mask}}=1.0$, $\lambda_{\text{part}}=0.5$, $\lambda_{\text{text}}=1.0$, $\lambda_{\text{cov}}=0.5$, $\lambda_{\text{overlap}}=0.1$, $\lambda_{\text{global}}=1.0$. We use AdamW with an initial learning rate of 3e-4 and cosine annealing to a minimum of 5e-6.

\begin{figure}[!htbp]
    \centering
    \includegraphics[width=\linewidth]{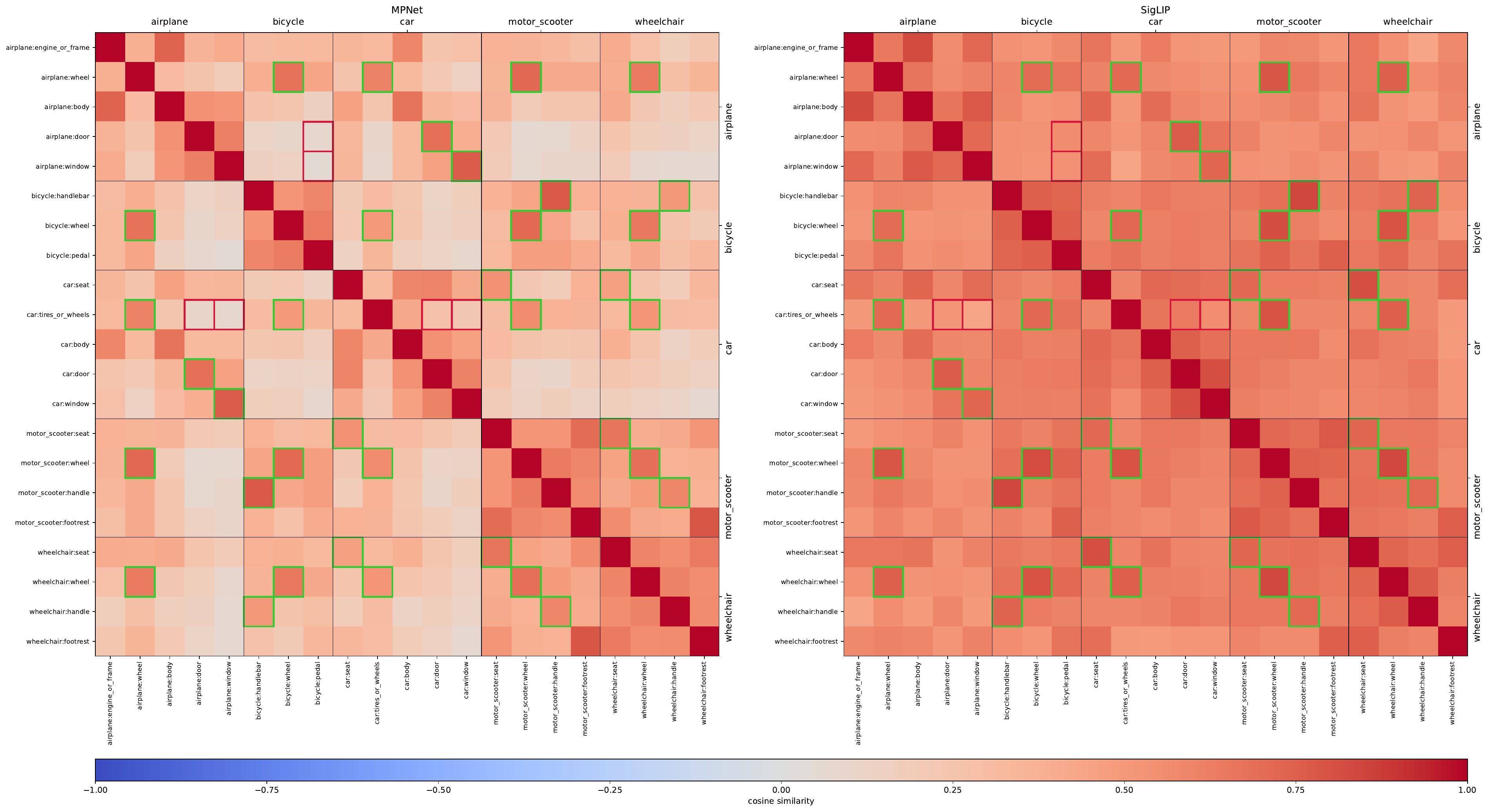}
    \caption{\small Pairwise cosine similarity heatmaps between
    text embeddings for MPNet (left) and SigLiP (right). (Zoom in for labels)}
    \label{fig:text-enc-comparison}
\end{figure}

\paragraph{Choice of Text Encoder} We adopt MPNet~\cite{song2020mpnet} as our text encoder for part descriptions rather than CLIP / SigLIP due to its superior structure-preserving properties for sentence-level embeddings. Standard vision-language models like SigLIP often suffer from representational collapse when applied to sub-object components. Because SigLIP is optimized for holistic scene descriptions, it fails to distinguish between semantically diverse part labels (e.g., those generated by LLMs like Gemini). This results in an undifferentiated embedding space that precludes the use of discriminative partlet-based learning. 
In \cref{fig:text-enc-comparison} for instance, MPNet correctly assigns high similarities (>0.8) to functionally equivalent parts across classes - such as wheels (airplane, car, bicycle, wheelchair), doors (airplane, car), and handles (scooter, bicycle, wheelchair), while maintaining low similarities (<0.3) between parts with different affordances, such as tires vs. doors/windows or pedals vs. airplane components. In contrast, SigLIP assigns uniformly high similarities to both sets, collapsing the semantic space and preventing our partlets from learning meaningful text-conditioned part alignment during training. 

\paragraph{Why affordance descriptions?} A key motivation for incorporating affordance information into part annotations is rooted in the cognitive science understanding that humans interpret and define object parts not just by geometry, but by their function, context, and description~\cite{tversky84}. Short or generic part names (e.g., ``leg'', ``handle'') are often ambiguous across different objects, lacking any semantic detail regarding the role or meaning of a part within a specific context. For example, ``legs'' fulfill distinct structural functions and take on different forms for chairs, tables, or sofas, a distinction that arises from their object-specific affordances. Prior work demonstrates that affordance-based cues and descriptive information guide human part recognition, reducing label ambiguity and supporting more robust reasoning and communication. Thus, by situating part annotations within functional and contextual descriptions, our approach enables higher-quality, less ambiguous labeling, consistent with cognitive models of human object understanding. \\

\paragraph{Datasets.} We train on 40,982 shapes from three datasets: 3DCoMPaT++ (8,627), PartNet (32,141), and Find3D (124). All use fine-grained part labels. For evaluation, we hold out 206 shapes: 126 objects (42 categories) from 3DCoMPaT++, 72 objects (24 categories) from PartNet, and 8 novel objects (8 categories) from Find3D.

\begin{figure*}[!h]
    \centering
    \includegraphics[width=\linewidth]{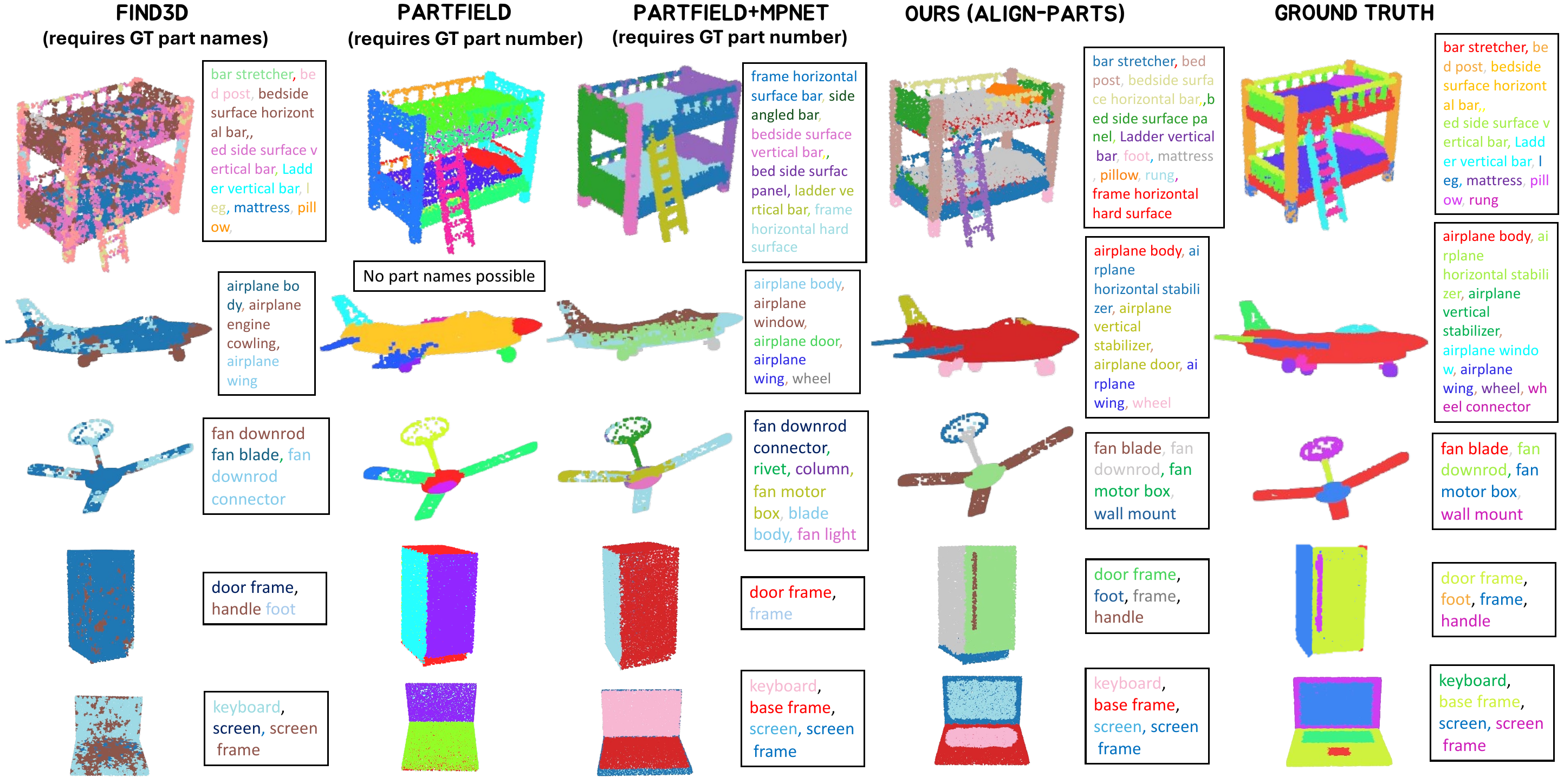}
\caption{\textbf{Qualitative Results.} ALIGN-Parts segments and names 3D parts robustly in a single feed-forward pass (rightmost column). Find3D~\cite{ma2025find} (first column) fails despite ground-truth part names, unable to segment the laptop in \cref{fig:results} (bottom row). PartField~\cite{liu2025partfield} (second column) also fails: it requires ground-truth part counts for clustering, missegments bed bunks (top row), and misses refrigerator handles (second-to-last row). Our strong baseline without Partlets (third column) exhibits similar errors. In contrast, ALIGN-Parts correctly segments tiny parts, such as handles, and groups semantically similar instances (e.g., all ceiling fan blades into a single cluster).}
\label{fig:results}

\end{figure*}
\section{Experiments}
\label{sec:exp}

In this section, we discuss our methodology for unifying 3D part benchmarks with inconsistent part naming conventions for the same object classes (\cref{subsec:vocab_comp}), introduce our baselines and metrics for the named 3D part segmentation task (\cref{subsec:comp_baselines}), and show qualitative and quantitative comparison of ALIGN-Parts with related baselines (\cref{subsec:results}). We ablate different components and inference modes of ALIGN-Parts in \cref{subsec:ablations} and \cref{subsec:inf_ablations}. We end the section with 2 applications of our method in \cref{subsec:label_transfer} and \cref{subsec:texparts}.

\subsection{3D Part Annotation Alignment \& Vocabulary Compression}
\label{subsec:vocab_comp}

Our 3D part annotation alignment and vocabulary compression methodology employs a two-stage pipeline that combines MPNet embeddings for candidate generation and Gemini LLM verification to reject spurious matches as well as identify and merge duplicate classes and parts across a unified 3D object taxonomy. 

For confirmed matches, the system successfully identifies semantically equivalent entities with high MPNet similarity scores that Gemini validates as identical: for example, "laptop\_computer" and "laptop" (similarity: 0.944) are merged because Gemini recognizes that "both candidate names refer to the exact same physical device ... consistently define it as a portable personal computer designed for mobile use". Similarly, \emph{microwave\_oven} (3DCoMPaT) and \emph{microwave} (PartNet) with similarity 0.902 are merged after Gemini confirmed they "describe the same kitchen appliance". Within the "microwave\_oven" class, "door\_glass" and "glass" (similarity: 0.865) are unified because Gemini concludes "both descriptions refer to the transparent panel integrated into the door ... that allows viewing food and contains radiation". The secondary part name 'glass' is a concise reference to the "door\_glass". Similarly, "bed\_footboard" and "footboard" (similarity: 0.953) are merged as Gemini states, " 'bed\_footboard' is a more explicit naming of 'footboard', and their descriptions are semantically identical, describing a panel at the foot of the bed opposite the headboard." 

For rejected pairs, the system correctly distinguishes semantically distinct parts despite high embedding similarity: "car\_front\_bumper" and "car\_rear\_bumper" (similarity: 0.879) are kept separate because Gemini determines "while both parts are bumpers with the same protective function, their specified locations (front vs. rear) make them distinct semantic parts for a 3D car object," and within the "chair" class, "back\_frame\_horizontal\_rod" and "back\_frame\_vertical\_rod" (similarity: 0.943) remain separate because Gemini explains "the parts are distinct based on their orientation within the back frame: one is explicitly described as a `horizontal rod' providing reinforcement for the backrest, while the other is a `vertical rod' providing structural support." The compressed vocabulary output maintains canonical names (choosing more verbose/descriptive variants), aggregates part counts across merged entities, and produces a mapping log that records every alias resolution for downstream lookup when legacy names are encountered during inference. This compressed vocabulary enables training with unified part semantics.

\subsection{Comparison for Named 3D Part Segmentation}
\label{subsec:comp_baselines}
\paragraph{Baselines.} While no prior work addresses \emph{named} 3D part segmentation end-to-end, we compare against two recent baselines: PartField~\cite{liu2025partfield} and Find3D~\cite{ma2025find}. PartField achieves state-of-the-art class-agnostic segmentation but cannot name parts. It clusters a learned feature field (requiring ground-truth part count K). Find3D matches per-point features to SigLIP embeddings of \emph{provided} part queries, needing the list of ground-truth part names as input. 

\paragraph{PartField+MPNet (our baseline w/o partlets).} We extend PartField with semantic alignment: a linear head maps PartField features to a shared space with MPNet text embeddings, trained via InfoNCE loss. However, it still requires predicting K via an auxiliary classifier, making it brittle to over- or under-segmentation errors in K prediction. Further architecture details are provided in \cref{sec:bridge}. 

\begin{figure}[!htbp]
    \centering
\includegraphics[width=\linewidth]{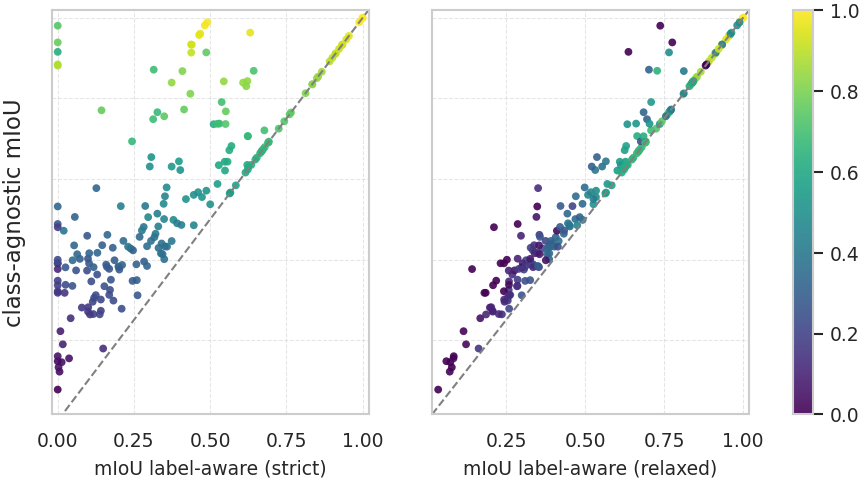}

\caption{\textbf{Proposed Metrics Correlation Analysis.} Correlation analysis between our proposed label-aware mIoU metrics (strict/relaxed) and class-agnostic mIoU, computed on segmentation results from our ALIGN-Parts model. The strict label-aware metric (left) shows moderate agreement with class-agnostic mIoU (Pearson $r=0.739$, Spearman $\rho=0.730$, $N=206$), while the relaxed variant (right) demonstrates near-perfect correlation (Pearson $r=0.978$, Spearman $\rho=0.974$, $N=206$). These findings indicate that our model achieves strong semantic and quantitative consistency, further supporting the use of the relaxed metric as a robust evaluation protocol for semantic 3D part segmentation.}
\label{fig:mioucorr}
\end{figure}

\begin{figure}[h!]
    \centering
\includegraphics[width=.8\linewidth]{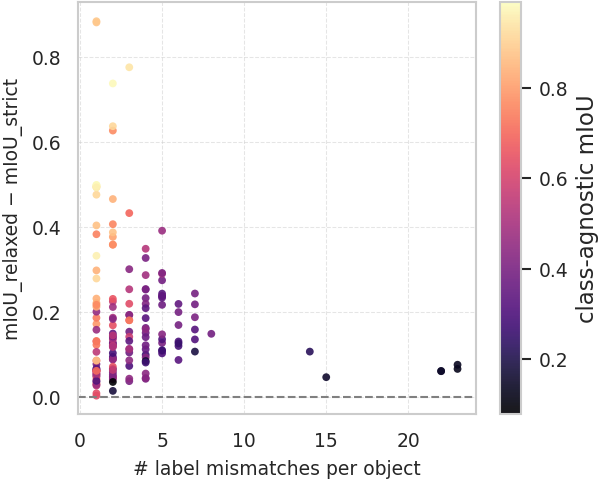}

\caption{\textbf{Label Mismatches and Metric Robustness.} Distribution of test objects according to the number of label mismatches (x-axis) and the resulting increase in mIoU from strict to relaxed label-aware matching (y-axis), colored by class-agnostic mIoU. Most objects with low mismatches exhibit small or moderate increases in mIoU, while objects with higher mismatch counts still do not show large outliers in metric difference, supporting the robustness of our evaluation protocol. The absence of extreme discrepancies suggests that the relaxed metric yields a stable and meaningful improvement, even for challenging cases, confirming its reliability for assessing semantic part segmentation on our test set.}
\label{fig:mioumismatch}
\end{figure}

\paragraph{Metrics.} We evaluate our method using three complementary metrics that progressively incorporate semantic label correctness. (1) Class-agnostic mIoU: Following prior work~\cite{liu2025partfield, yang2024sampart3d}, for each ground-truth part, we compute the maximum IoU across all predicted segments and average these values, ignoring semantic labels entirely - this captures pure geometric segmentation quality. (2) Label-Aware mIoU (strict, LA-mIoU): For each ground-truth part, we identify the predicted segment with the highest geometric overlap (as in class-agnostic mIoU), then assign credit only if its semantic label exactly matches the ground truth; otherwise, the part contributes 0.0 - this measures joint geometry-semantic accuracy with strict label matching. (3) Relaxed Label-Aware mIoU (rLA-mIoU): Identical segment selection to strict LA-mIoU, but instead of binary label matching, we weight the IoU by the cosine similarity between MPNet text embeddings of predicted and ground-truth labels, giving partial credit to semantically related predictions (e.g., "screen" vs. "monitor") - this captures semantic near-misses that strict matching penalizes. By construction, we have class-agnostic mIoU $\geq$ rLA-mIoU $\geq$ LA-mIoU (strict), where equality holds only when all predictions have perfect label agreement. The gap between class-agnostic and label-aware metrics reveals semantic prediction errors, while the gap between strict and relaxed variants quantifies label confusion on semantically similar parts. Correlation plots (\cref{fig:mioucorr} and \cref{fig:mioumismatch}) empirically validate that relaxed scoring recovers significant semantic credit on near-miss predictions.

\subsection{Results}
\label{subsec:results}

\begin{table*}[t]
  \centering
  \small
  \caption{Evaluation of ALIGN-Parts and related baselines on our test set. In addition to the usual mIoU metric for evaluating class-agnostic part segmentation, we introduce 2 metrics more suited for our named part-segmentation task, namely, label-aware mIoU (LA-mIoU) and relaxed label-aware mIoU (rLA-mIoU).}
  \label{tab:main_tab}
  \setlength{\tabcolsep}{6pt}
  \resizebox{\textwidth}{!}{%
  \begin{tabular}{lcccccccccccc}
    \toprule
    \multirow{2}{*}{Variant} & \multicolumn{3}{c}{3DCoMPaT} & \multicolumn{3}{c}{Find3D} & \multicolumn{3}{c}{PartNet} & \multicolumn{3}{c}{Average} \\
    \cmidrule(lr){2-4} \cmidrule(lr){5-7} \cmidrule(lr){8-10} \cmidrule(lr){11-13}
      & mIoU$\uparrow$ & LA-mIoU$\uparrow$ & rLA-mIoU$\uparrow$ & mIoU$\uparrow$ & LA-mIoU$\uparrow$ & rLA-mIoU$\uparrow$ & mIoU$\uparrow$ & LA-mIoU$\uparrow$ & rLA-mIoU$\uparrow$ & mIoU$\uparrow$ & LA-mIoU$\uparrow$ & rLA-mIoU$\uparrow$ \\
    \midrule
    PartField & \cellcolor{tabsecond}0.371 & n/a & n/a & \cellcolor{tabfirst}0.662 & n/a & n/a & \cellcolor{tabsecond}0.521 & n/a & n/a & \cellcolor{tabsecond}0.518 & n/a & n/a \\
    Find3D & 0.239 & \cellcolor{tabthird}0.072 & \cellcolor{tabthird}0.178 & 0.379 & \cellcolor{tabfirst}0.232 & \cellcolor{tabthird}0.324 & 0.354 & \cellcolor{tabthird}0.204 & \cellcolor{tabthird}0.298 & 0.324 & \cellcolor{tabthird}0.169 & \cellcolor{tabthird}0.267 \\
    PartField+MPNet & \cellcolor{tabthird}0.316 & \cellcolor{tabsecond}0.185 & \cellcolor{tabsecond}0.259 & \cellcolor{tabthird}0.590 & \cellcolor{tabsecond}0.137 & \cellcolor{tabsecond}0.451 & \cellcolor{tabthird}0.446 & \cellcolor{tabsecond}0.276 & \cellcolor{tabsecond}0.394 & \cellcolor{tabthird}0.451 & \cellcolor{tabsecond}0.199 & \cellcolor{tabsecond}0.368 \\
    ALIGN-Parts & \cellcolor{tabfirst}0.453 & \cellcolor{tabfirst}0.268 & \cellcolor{tabfirst}0.391 & \cellcolor{tabsecond}0.595 & \cellcolor{tabthird}0.133 & \cellcolor{tabfirst}0.466 & \cellcolor{tabfirst}0.753 & \cellcolor{tabfirst}0.546 & \cellcolor{tabfirst}0.729 & \cellcolor{tabfirst}0.600 & \cellcolor{tabfirst}0.316 & \cellcolor{tabfirst}0.529 \\
    \bottomrule
  \end{tabular}
  }
\end{table*}

\begin{figure}[htp!]
    \centering
\includegraphics[width=\linewidth]{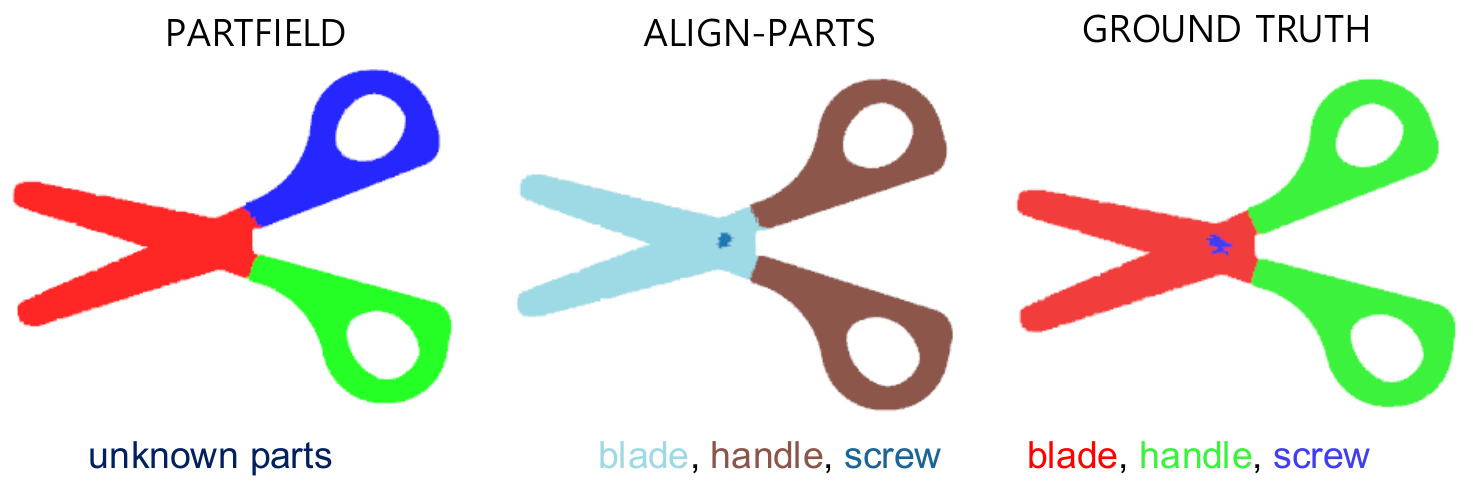}

    \caption{\small ALIGN-Parts correctly segments the tiny screw despite training with sparser points (10k vs. 100k).}
    \label{fig:fineparts}
    
\end{figure}

In \cref{fig:results} and \cref{fig:fineparts} and \cref{tab:main_tab}, we show that ALIGN-Parts comprehensively outperforms all baselines on both class-agnostic segmentation (mIoU) and named part segmentation (LA-mIoU). On average mIoU, we outperform PartField by 15.8\%, whereas on LA-mIoU and rLA-mIoU, we improve over PartField+MPNet by 58.8\% and 43.8\%, respectively, while being $100\times$ faster because we do not require running a K-means clustering algorithm.

Qualitatively, our baselines show several weaknesses. PartField often fragments instances of the same part into multiple segments after clustering, contradicting human labeling conventions in 3DCoMPaT++ and PartNet. This likely stems from its use of SAM~\cite{kirillov2023segany} to extract unlabeled parts from 2D renderings. In contrast, ALIGN-Parts, trained on human annotations, correctly groups instances: e.g., all four bed posts of a double bed and all three wheels of an airplane are identified as semantically single parts (\cref{fig:fineparts}).

Find3D learns per-point semantic vectors without considering shape geometry, resulting in noisy and overlapping segmentations. PartField+MPNet's reliance on predicted cluster counts leads to under-segmentation, as fine parts such as the refrigerator handle and laptop screen frame are missed (\cref{fig:fineparts}). Both Find3D and PartField often fail to segment relatively simple structures, such as fan blades, whereas ALIGN-Parts accurately segments and annotates fine-grained parts in complex shapes.

\paragraph{Runtime Comparison.} In terms of runtime, our method compares favorably against all baselines. Find3D runs in $\approx 0.25s$, whereas both PartField and PartField+MPNet require $\approx 4s$, where the majority of the runtime is consumed by K-means clustering. ALIGN-Parts, being a one-shot feedforward method, needs only $\approx 0.05s$ (barring feature pre-processing) to produce labelled 3D parts.

\begin{table*}[t]
  \centering
  \small
  \caption{Ablation study on 3DCoMPaT, Find3D, and PartNet. We report mIoU, label-aware mIoU (LA-mIoU), and relaxed label-aware mIoU (rLA-mIoU).}
  \label{tab:ablation}
  \setlength{\tabcolsep}{6pt}
  \resizebox{\textwidth}{!}{%
  \begin{tabular}{lcccccccccccc}
    \toprule
    \multirow{2}{*}{Variant} & \multicolumn{3}{c}{3DCoMPaT} & \multicolumn{3}{c}{Find3D} & \multicolumn{3}{c}{PartNet} & \multicolumn{3}{c}{Average} \\
    \cmidrule(lr){2-4} \cmidrule(lr){5-7} \cmidrule(lr){8-10} \cmidrule(lr){11-13}
      & mIoU$\uparrow$ & LA-mIoU$\uparrow$ & rLA-mIoU$\uparrow$ & mIoU$\uparrow$ & LA-mIoU$\uparrow$ & rLA-mIoU$\uparrow$ & mIoU$\uparrow$ & LA-mIoU$\uparrow$ & rLA-mIoU$\uparrow$ & mIoU$\uparrow$ & LA-mIoU$\uparrow$ & rLA-mIoU$\uparrow$ \\
    \midrule
    Base Model & 0.228 & 0.021 & 0.141 & 0.374 & 0.041 & 0.239 & 0.335 & 0.027 & 0.201 & 0.312 & 0.030 & 0.194 \\
    No $\mathcal{L}_{\text{cov}}$, $\mathcal{L}_{\text{ov}}$, $\mathcal{L}_{\text{txt}}$ & 0.233 & 0.019 & 0.140 & 0.382 & 0.003 & 0.198 & 0.357 & 0.041 & 0.223 & 0.324 & 0.021 & 0.187 \\
    No $\mathcal{L}_{\text{cov}}$, $\mathcal{L}_{\text{ov}}$ & \cellcolor{tabsecond}0.422 & \cellcolor{tabsecond}0.193 & \cellcolor{tabsecond}0.338 & \cellcolor{tabthird}0.499 & \cellcolor{tabthird}0.110 & \cellcolor{tabthird}0.384 & \cellcolor{tabsecond}0.664 & \cellcolor{tabsecond}0.414 & \cellcolor{tabsecond}0.609 & \cellcolor{tabsecond}0.528 & \cellcolor{tabsecond}0.239 & \cellcolor{tabsecond}0.443 \\
    Geo Input Only & 0.224 & 0.014 & 0.134 & 0.384 & 0.027 & 0.187 & 0.332 & 0.040 & 0.210 & 0.313 & 0.027 & 0.177 \\
    Feature Concat & 0.221 & 0.017 & 0.134 & 0.367 & 0.058 & 0.239 & 0.317 & 0.033 & 0.194 & 0.302 & 0.036 & 0.189 \\
    PartField+MPNet & \cellcolor{tabthird}0.316 & \cellcolor{tabthird}0.185 & \cellcolor{tabthird}0.259 & \cellcolor{tabsecond}0.590 & \cellcolor{tabfirst}0.137 & \cellcolor{tabsecond}0.451 & \cellcolor{tabthird}0.446 & \cellcolor{tabthird}0.276 & \cellcolor{tabthird}0.394 & \cellcolor{tabthird}0.451 & \cellcolor{tabthird}0.199 & \cellcolor{tabthird}0.368 \\
    ALIGN-Parts & \cellcolor{tabfirst}0.453 & \cellcolor{tabfirst}0.268 & \cellcolor{tabfirst}0.391 & \cellcolor{tabfirst}0.595 & \cellcolor{tabsecond}0.133 & \cellcolor{tabfirst}0.466 & \cellcolor{tabfirst}0.753 & \cellcolor{tabfirst}0.546 & \cellcolor{tabfirst}0.729 & \cellcolor{tabfirst}0.600 & \cellcolor{tabfirst}0.316 & \cellcolor{tabfirst}0.529 \\
    \bottomrule
  \end{tabular}
  }
\end{table*}

\paragraph{Fine-Part Localization}
Despite using only 10k sampled points, ALIGN-Parts segments fine parts, such as the \emph{screw} of scissors (\cref{fig:fineparts}) - structures that PartField, trained with 100k points, cannot localize.

\begin{figure*}[t!]
    \centering
    \includegraphics[width=0.98\linewidth]{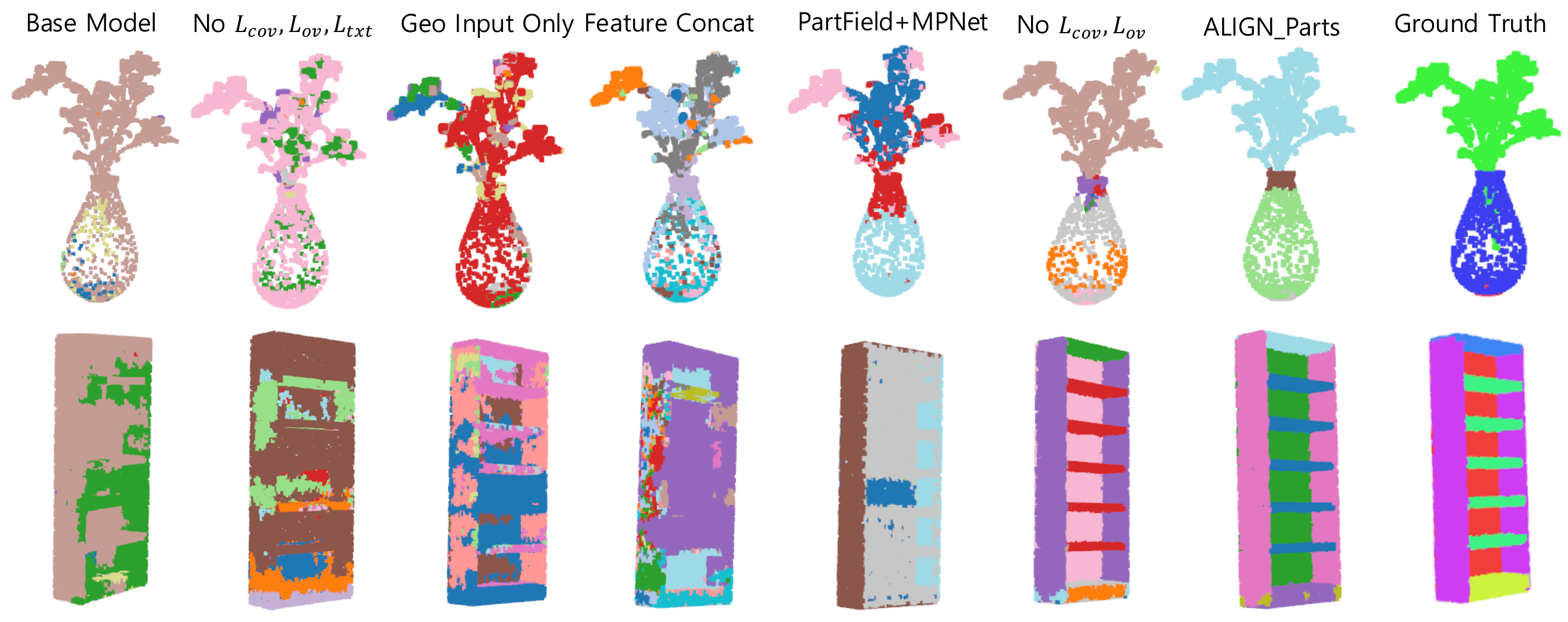} %
\caption{\textbf{Ablation (Qualitative results).}  Results improve from left to right as components are added sequentially, with ground truth in the final column. The first five columns result in significant misalignments and segmentation leakage. Neither geometric-only features nor naive DINO concatenation improves performance. Major gains arise from Partlets (see sixth and seventh columns) and the coverage loss, which refine fine details and are consistent with quantitative metrics in \cref{tab:ablation}.}
\label{fig:ablation}
\end{figure*}

\subsection{Ablations}
\label{subsec:ablations}

\cref{tab:ablation} and \cref{fig:ablation} evaluate ALIGN-Parts's design choices, showing quantitative and qualitative improvements from each component.

\paragraph{Baseline Comparisons.} Using only PartField geometric features, our base model ($\mathcal{L}_{\text{mask}}, \mathcal{L}_{\text{part}}, \mathcal{L}_{\text{global}}$) performs slightly better than naive concatenation of PartField and DINOv2 features, indicating that simple multi-modal fusion can be counterproductive. A variant using raw geometry without PartField shows no improvement, confirming that learned geometric features are essential.

\paragraph{Progressive Component Addition.} Adding DINOv2 appearance features yields modest gains. Incorporating the InfoNCE text alignment loss ($\mathcal{L}_{\text{txt}}$) significantly improves both LA-mIoU and rLA-mIoU; this variant is the only one, besides our full model, that correctly segments fine-grained parts, such as plants in vases or shelves in storage furniture. Finally, adding auxiliary regularizers ($\mathcal{L}_{\text{cov}}, \mathcal{L}_{\text{overlap}}$) yields the complete ALIGN-Parts model, which achieves peak performance both qualitatively and quantitatively.

\subsection{Inference-time Ablations}
\label{subsec:inf_ablations}

\begin{table*}[t!]
  \centering
  \caption{Evaluation of different inference modes of ALIGN-Parts on our test set, using mean IoU (mIoU), label-aware mIoU (LA-mIoU), and relaxed label-aware mIoU (rLA-mIoU). Providing additional ground-truth part count information only slightly improves the model's performance on rLA-mIoU showing that ALIGN-Parts often estimates accurate part cardinality based on just the input geometric and appearance features of a 3D shape.}
  \setlength{\tabcolsep}{4pt}
  \resizebox{\textwidth}{!}{%
  \begin{tabular}{lcccccccccccc}
    \toprule
    \multirow{2}{*}{Variant} & \multicolumn{3}{c}{3DCoMPaT (126)} & \multicolumn{3}{c}{Find3D (8)} & \multicolumn{3}{c}{PartNet (72)} & \multicolumn{3}{c}{Average} \\
    \cmidrule(lr){2-4} \cmidrule(lr){5-7} \cmidrule(lr){8-10} \cmidrule(lr){11-13}
      & mIoU$\uparrow$ & LA-mIoU$\uparrow$ & rLA-mIoU$\uparrow$ & mIoU$\uparrow$ & LA-mIoU$\uparrow$ & rLA-mIoU$\uparrow$ & mIoU$\uparrow$ & LA-mIoU$\uparrow$ & rLA-mIoU$\uparrow$ & mIoU$\uparrow$ & LA-mIoU$\uparrow$ & rLA-mIoU$\uparrow$ \\
    \midrule
    \shortstack{Clustering +\\ Part Number} & \cellcolor{tabthird}0.370 & n/a & n/a & \cellcolor{tabthird}0.528 & n/a & n/a & \cellcolor{tabthird}0.537 & n/a & n/a & \cellcolor{tabthird}0.478 & n/a & n/a \\
    +Part Number & \cellcolor{tabsecond}0.452 & \cellcolor{tabfirst}0.268 & \cellcolor{tabsecond}0.389 & \cellcolor{tabfirst}0.625 & \cellcolor{tabfirst}0.138 & \cellcolor{tabfirst}0.473 & \cellcolor{tabfirst}0.757 & \cellcolor{tabfirst}0.559 & \cellcolor{tabfirst}0.737 & \cellcolor{tabfirst}0.611 & \cellcolor{tabfirst}0.322 & \cellcolor{tabfirst}0.533 \\
    No Part Number & \cellcolor{tabfirst}0.453 & \cellcolor{tabsecond}0.268 & \cellcolor{tabfirst}0.391 & \cellcolor{tabsecond}0.595 & \cellcolor{tabsecond}0.133 & \cellcolor{tabsecond}0.466 & \cellcolor{tabsecond}0.753 & \cellcolor{tabsecond}0.546 & \cellcolor{tabsecond}0.729 & \cellcolor{tabsecond}0.600 & \cellcolor{tabsecond}0.316 & \cellcolor{tabsecond}0.529 \\
    \bottomrule
  \end{tabular}
  }
  \label{tab:inf_ablation}
\end{table*}

\begin{figure*}[t]
\centering
\includegraphics[width=0.9\linewidth]{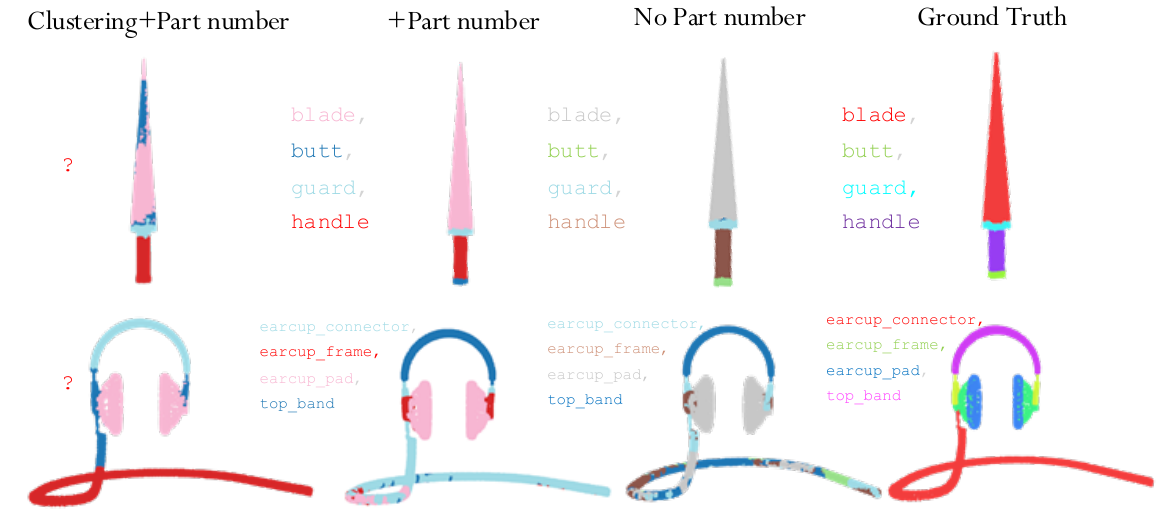}
\caption{Inference-time Ablations. We assess two additional inference modes in our ALIGN-Parts model. The \emph{Clustering+Part number} mode disregards text vocabularies and applies k-means to the fused geometric and appearance features, generating class-agnostic clusters and probing how well raw feature representations alone support meaningful part segmentation. The \emph{+Part number} mode uses the true part count for each shape, activating exactly $M$ partlets with the highest saliency scores - computed by combining partness confidence and average mask coverage - and assigning every point to its best matching mask; this tests whether supplying ground-truth cardinality adds value compared to dynamic, data-driven part discovery. Both setups are compared to the default dynamic activation (No Part number) and the ground truth, highlighting that ALIGN-Parts robustly estimates part cardinality and produces accurate segmentations even without explicit part label or count supervision.}
\label{fig:inferenceablation}
\end{figure*}

We evaluate two additional inference modes of our ALIGN-Parts model, extending beyond the primary dynamic part activation approach to better understand the contributions of part cardinality and label information in our segmentation pipeline. The first alternative mode, which we term the \emph{clustering+part number} setting, completely forgoes the use of any part vocabularies or text labels during inference. Instead, it relies solely on the fused geometric and appearance features output by the model, upon which we run k-means clustering to produce purely class-agnostic instance clusters. This setup rigorously probes the ability of the learned feature embeddings, untethered to semantic labels, to support coherent part decompositions across diverse objects, essentially isolating the impact of visual and geometric cues alone. The second mode, called \emph{+Part number}, examines whether providing the model with the exact ground-truth part count for each input shape improves segmentation quality compared to the default setting, where the model dynamically infers the number of parts to activate. After producing all candidate partlet masks and calculating their partness scores, this mode ranks the partlets by a saliency score, which is a composite measure combining the confidence that a partlet corresponds to an actual part (i.e., partness) and the average mask coverage over the point cloud (mean mask probability mass over points). From this ranking, the top $M$ partlets are retained, where $M$ is the true number of parts for the target shape, and every point in the shape is assigned the best matching mask among these selected partlets to yield a hard $K$-way partition. These inference ablation modes and their qualitative and quantitative outcomes are detailed and visualized in \cref{fig:inferenceablation} and \cref{tab:inf_ablation}, demonstrating that ALIGN-Parts is able to robustly estimate accurate part cardinality and segmentation even without explicit part label or count guidance, and that the fused multimodal features alone provide meaningful cues towards coherent part delineation. This analysis not only highlights the flexibility and robustness of ALIGN-Parts at inference time, but also emphasizes the benefits of its design choices in learning and leveraging rich feature representations supporting both semantic and instance-aware part segmentation.

\subsection{TexParts Dataset}
\label{subsec:texparts}

\begin{figure*}[!htbp]
    \centering
    \includegraphics[width=\linewidth]{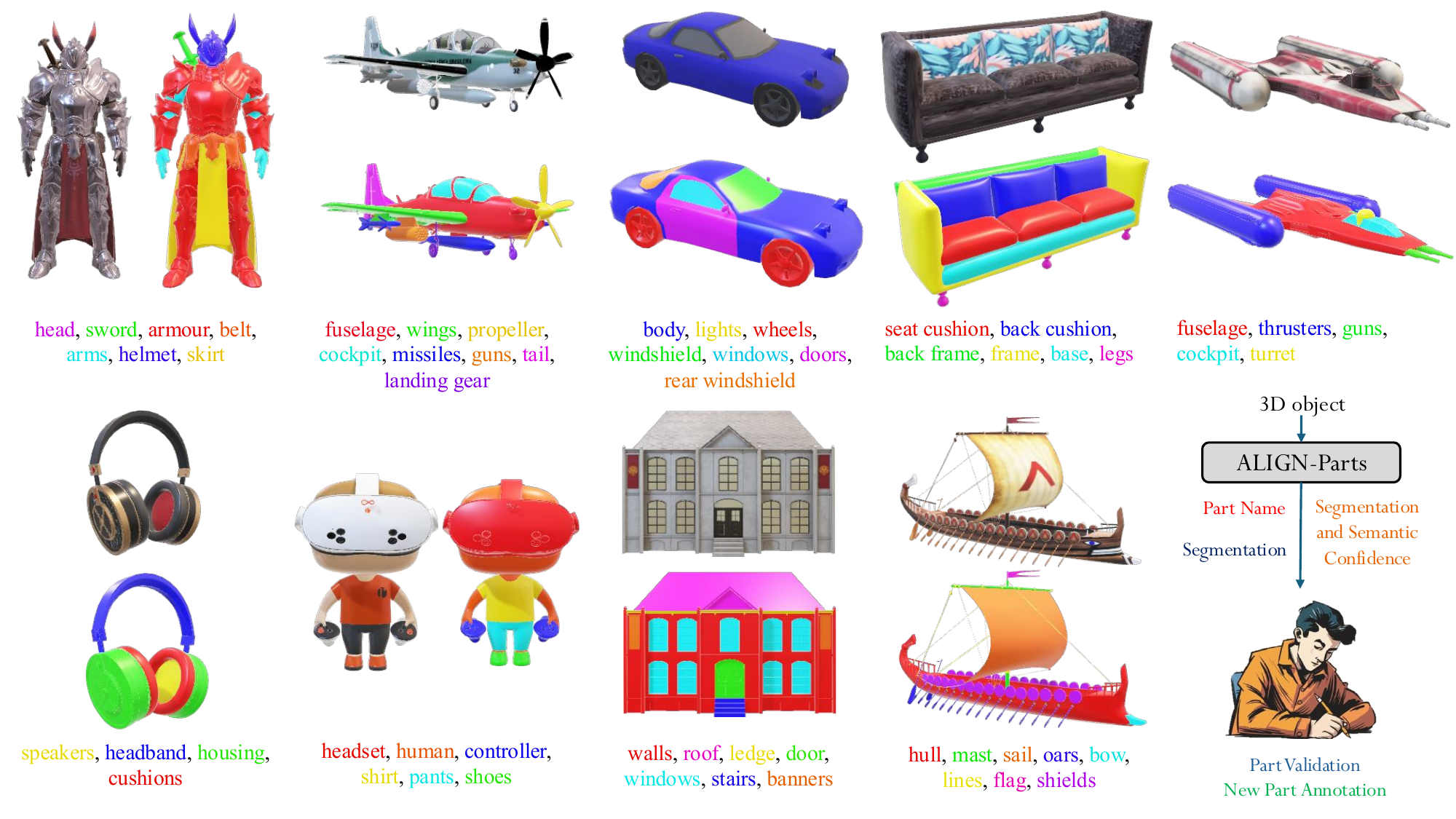}
\caption{\textbf{TexParts Dataset.} We demonstrate human-in-the-loop annotation of Texverse~\cite{zhang2025texverse} using ALIGN-Parts, enabling scalable dense 3D part segmentation.}
\label{fig:texpartssup}
\end{figure*}

A central aim of our approach is to enable the construction of a high-quality 3D part annotation dataset with minimal manual intervention, ensuring both unified and comprehensive part labeling at scale. For this purpose, we select the TexVerse dataset as our unannotated 3D source corpus, leveraging its exceptional quality, high-resolution textures, and extensive diversity of 3D assets~\cite{zhang2025texverse}. TexVerse consists of over 850,000 unique 3D models with physically based rendering (PBR) materials and rich metadata, making it an ideal foundation for large-scale part segmentation.

Our pipeline begins with the automated filtering of TexVerse models: using Gemini-Flash LLM, we combine thumbnail images and other metadata to preselect high-quality objects and exclude inadequate or malformed models. Next, we apply our ALIGN-Parts model and save, for each shape, its predicted part masks, part names, and both semantic and segmentation confidence scores. To prioritize downstream annotation effort, we sort objects by their average confidence score (in descending order) so that annotators see the most reliable candidates first. Selected objects are then routed to human annotators for validation and correction.

During annotation, annotators use several aids: a part name prompting tool for searching or extending the active part vocabulary, and (optionally) the ability to reference unlabeled geometric part masks generated by PartField. Our annotation process is explicitly bilevel; Phase One focuses on validating and making minor edits to ALIGN-Parts predictions, while Phase Two addresses new or missing parts that may require more substantive manual annotation. By the time of submission of this work, the first phase had covered approximately 8,000 objects, comprising around 14,000 unique part categories. Examples from the current dataset are shown in \cref{fig:texpartssup}. 

A key observation from our annotation workflow is the dramatic reduction in manual effort enabled by our methodology: annotating 3D objects from scratch typically takes anywhere from 15 to 25 minutes per shape, while our model-assisted pipeline reduces annotation time to just 3 to 5 minutes on average - a time saving of approximately 5--8$\times$ without sacrificing annotation quality. 

Importantly, and in clear contrast to recent approaches that keep their Objaverse-derived part annotations closed-source, we commit to releasing TexParts as a public resource upon publication, with the aim of advancing large-scale open research in semantic 3D part understanding.

\section{Limitations and Future Work}
The primary limitations of our work stem from the relatively restricted set of objects and parts on which ALIGN-Parts has been trained, compared to the vast (though finite) variety of parts that occur in the real world. This gap is largely due to the scarcity of large-scale 3D datasets with dense part annotations and a unified, operational definition of what constitutes a ``part''. In effect, this creates a chicken-and-egg problem: ALIGN-Parts was designed to enable robust 3D part annotation at scale, yet the robustness and coverage of the model itself are constrained by the limited annotated data available for training.

Future work will focus on mitigating this dependency by exploring self-supervised or weakly supervised formulations and by incorporating stronger 3D priors, for example from generative models or skeletal/medial representations. Another important direction is to reduce the current reliance on frozen PartField features by enabling full end-to-end training of the geometric feature extractor, which was not pursued here primarily due to computational constraints rather than methodological ones. Despite these issues, our framework is immediately usable by parties with abundant compute and proprietary 3D assets, who can scale ALIGN-Parts to richer, closed-source datasets and drive progress towards truly large-scale 3D scene understanding at the part level.

\section{Discussion}
ALIGN-Parts reframes \textit{semantic} 3D part segmentation as a set alignment problem, where Partlets trained via optimal transport jointly learn geometry and semantics without part count supervision. Unlike Find3D's per-point alignments or PartField's brittle clustering, our end-to-end matching produces fast and coherent named parts directly (\cref{fig:results}). LLM-generated affordance descriptions ("door handle is grasped to open a door") are important for disambiguating fine-grained parts that confuse simple part names, and MPNet's handling of long-form text outperforms CLIP/SigLIP (\cref{fig:text-enc-comparison}). Despite using 10× fewer points than PartField, we achieve superior fine-part localization (e.g., a scissors screw, \cref{fig:fineparts}), confirming that semantic part-level representations are more data-efficient than dense per-point features. Key limitations are: noisy real-world scans challenge our manifold assumptions, Mahalanobis confidence degrades under distribution shift, and open-vocabulary generalization is limited to categories similar to the training data. Future work should extend this to articulated objects and integrate part-level alignments into foundation 3D models for manipulation and generation. By bridging dense geometry and structured language, ALIGN-Parts enables scalable, semantically rich 3D asset creation.

\clearpage
{
    \small
    \bibliographystyle{ieeenat_fullname}
    \bibliography{main}

@String(CVPR= {IEEE Conf. Comput. Vis. Pattern Recog.})

@String(ECCV= {Eur. Conf. Comput. Vis.})

@String(TOG= {ACM Trans. Graph.})

@article{vaswani2017attention,
  title={Attention is all you need},
  author={Vaswani, Ashish and Shazeer, Noam and Parmar, Niki and Uszkoreit, Jakob and Jones, Llion and Gomez, Aidan N and Kaiser, {\L}ukasz and Polosukhin, Illia},
  journal={Advances in neural information processing systems},
  volume={30},
  year={2017}
}

@article{kirillov2023segany,
  title={Segment Anything},
  author={Kirillov, Alexander and Mintun, Eric and Ravi, Nikhila and Mao, Hanzi and Rolland, Chloe and Gustafson, Laura and Xiao, Tete and Whitehead, Spencer and Berg, Alexander C. and Lo, Wan-Yen and Doll{\'a}r, Piotr and Girshick, Ross},
  journal={arXiv:2304.02643},
  year={2023}
}

@inproceedings{umam2024partdistill,
  title={Partdistill: 3d shape part segmentation by vision-language model distillation},
  author={Umam, Ardian and Yang, Cheng-Kun and Chen, Min-Hung and Chuang, Jen-Hui and Lin, Yen-Yu},
  booktitle={Proceedings of the IEEE/CVF Conference on Computer Vision and Pattern Recognition},
  pages={3470--3479},
  year={2024}
}

@article{slim20253dcompat++,
  title={3dcompat++: An improved large-scale 3d vision dataset for compositional recognition},
  author={Slim, Habib and Li, Xiang and Li, Yuchen and Ahmed, Mahmoud and Ayman, Mohamed and Upadhyay, Ujjwal and Abdelreheem, Ahmed and Prajapati, Arpit and Pothigara, Suhail and Wonka, Peter and others},
  journal={IEEE Transactions on Pattern Analysis and Machine Intelligence},
  year={2025},
  publisher={IEEE}
}

@article{song2020mpnet,
  title={Mpnet: Masked and permuted pre-training for language understanding},
  author={Song, Kaitao and Tan, Xu and Qin, Tao and Lu, Jianfeng and Liu, Tie-Yan},
  journal={Advances in neural information processing systems},
  volume={33},
  pages={16857--16867},
  year={2020}
}

@article{comanici2025gemini,
  title={Gemini 2.5: Pushing the frontier with advanced reasoning, multimodality, long context, and next generation agentic capabilities},
  author={Comanici, Gheorghe and Bieber, Eric and Schaekermann, Mike and Pasupat, Ice and Sachdeva, Noveen and Dhillon, Inderjit and Blistein, Marcel and Ram, Ori and Zhang, Dan and Rosen, Evan and others},
  journal={arXiv preprint arXiv:2507.06261},
  year={2025}
}

@article{zhang2025texverse,
  title={Texverse: A universe of 3d objects with high-resolution textures},
  author={Zhang, Yibo and Zhang, Li and Ma, Rui and Cao, Nan},
  journal={arXiv preprint arXiv:2508.10868},
  year={2025}
}

@inproceedings{koo2022partglot,
    title={PartGlot: Learning Shape Part Segmentation from Language Reference Games},
    author={
        Koo, Juil and
        Huang, Ian and
        Achlioptas, Panos and
        Guibas, Leonidas J and
        Sung, Minhyuk
    },
    booktitle={Proceedings of IEEE Conference on Computer Vision and Pattern Recognition (CVPR)},
    year={2022}
}

@inproceedings{li20223d_compat,
  title        = {{3DCoMPaT}: Composition of Materials on Parts of {3D} Things},
  author       = {Li, Yuchen and Upadhyay, Ujjwal and Slim, Habib and Abdelreheem, Ahmed and Prajapati, Arpit and Pothigara, Suhail and Wonka, Peter and Elhoseiny, Mohamed},
  booktitle    = {Proceedings of the European Conference on Computer Vision (ECCV)},
  year         = {2022}
}

@InProceedings{Mo_2019_CVPR,
    author = {Mo, Kaichun and Zhu, Shilin and Chang, Angel X. and Yi, Li and Tripathi, Subarna and Guibas, Leonidas J. and Su, Hao},
    title = {{PartNet}: A Large-Scale Benchmark for Fine-Grained and Hierarchical Part-Level {3D} Object Understanding},
    booktitle = {The IEEE Conference on Computer Vision and Pattern Recognition (CVPR)},
    month = {June},
    year = {2019}
}

@INPROCEEDINGS {Kalogerakis2017,
author = {Kalogerakis, Evangelos and Averkiou, Melinos and Maji, Subhransu and Chaudhuri, Siddhartha },
booktitle = { 2017 IEEE Conference on Computer Vision and Pattern Recognition (CVPR) },
title = {{ 3D Shape Segmentation with Projective Convolutional Networks }},
year = {2017},
volume = {},
ISSN = {1063-6919},
pages = {6630-6639},
doi = {10.1109/CVPR.2017.702},
url = {https://doi.ieeecomputersociety.org/10.1109/CVPR.2017.702},
publisher = {IEEE Computer Society},
address = {Los Alamitos, CA, USA},
month =Jul}

@article{Kalogerakis:2010:labelMeshes,
  Author    = {Evangelos Kalogerakis and Aaron Hertzmann and Karan Singh},
  Title     = {{L}earning {3}{D} {M}esh {S}egmentation and {L}abeling},
  Journal   = {ACM Transactions on Graphics},
  Volume    = {29},
  Number    = {3},
  Year = {2010},
}

@ARTICLE{7738348,
  author={Crouse, David F.},
  journal={IEEE Transactions on Aerospace and Electronic Systems}, 
  title={On implementing 2D rectangular assignment algorithms}, 
  year={2016},
  volume={52},
  number={4},
  pages={1679-1696},
  doi={10.1109/TAES.2016.140952}}

@inproceedings{kim2024partstad,
    title={PartSTAD: 2D-to-3D Part Segmentation Task Adaptation}, 
    author={Kim, Hyunjin and Sung, Minhyuk},
    booktitle={ECCV},
    year={2024}
}

@inproceedings{carion2020end,
  title={End-to-end object detection with transformers},
  author={Carion, Nicolas and Massa, Francisco and Synnaeve, Gabriel and Usunier, Nicolas and Kirillov, Alexander and Zagoruyko, Sergey},
  booktitle={European conference on computer vision},
  pages={213--229},
  year={2020},
  organization={Springer}
}

@misc{lin2025partcrafter,
  title={PartCrafter: Structured 3D Mesh Generation via Compositional Latent Diffusion Transformers}, 
  author={Yuchen Lin and Chenguo Lin and Panwang Pan and Honglei Yan and Yiqiang Feng and Yadong Mu and Katerina Fragkiadaki},
  year={2025},
  eprint={2506.05573},
  url={https://arxiv.org/abs/2506.05573}
}

@inproceedings{ma2025find,
  title={Find any part in 3d},
  author={Ma, Ziqi and Yue, Yisong and Gkioxari, Georgia},
  booktitle={Proceedings of the IEEE/CVF International Conference on Computer Vision},
  pages={7818--7827},
  year={2025}
}

@article{yang2025holopart,
  title={HoloPart: Generative 3D Part Amodal Segmentation}, 
  author={Yang, Yunhan and Guo, Yuan-Chen and Huang, Yukun and Zou, Zi-Xin and Yu, Zhipeng and Li, Yangguang and Cao, Yan-Pei and Liu, Xihui},
  journal={arXiv preprint arXiv:2504.07943},
  year={2025}
}

@inproceedings{dutt2024diffusion,
  title={Diffusion 3d features (diff3f): Decorating untextured shapes with distilled semantic features},
  author={Dutt, Niladri Shekhar and Muralikrishnan, Sanjeev and Mitra, Niloy J},
  booktitle={Proceedings of the IEEE/CVF Conference on Computer Vision and Pattern Recognition},
  pages={4494--4504},
  year={2024}
}

@inproceedings{liu2025partfield,
  title={Partfield: Learning 3d feature fields for part segmentation and beyond},
  author={Liu, Minghua and Uy, Mikaela Angelina and Xiang, Donglai and Su, Hao and Fidler, Sanja and Sharp, Nicholas and Gao, Jun},
  booktitle={Proceedings of the IEEE/CVF International Conference on Computer Vision},
  pages={9704--9715},
  year={2025}
}

@article{tang2024segment,
  title={Segment Any Mesh},
  author={Tang, George and Zhao, William and Ford, Logan and Benhaim, David and Zhang, Paul},
  journal={arXiv preprint arXiv:2408.13679},
  year={2024}
}

@article{yang2024sampart3d,
  title={Sampart3d: Segment any part in 3d objects},
  author={Yang, Yunhan and Huang, Yukun and Guo, Yuan-Chen and Lu, Liangjun and Wu, Xiaoyang and Lam, Edmund Y and Cao, Yan-Pei and Liu, Xihui},
  journal={arXiv preprint arXiv:2411.07184},
  year={2024}
}

@inproceedings{liu2023partslip,
  title={Partslip: Low-shot part segmentation for 3d point clouds via pretrained image-language models},
  author={Liu, Minghua and Zhu, Yinhao and Cai, Hong and Han, Shizhong and Ling, Zhan and Porikli, Fatih and Su, Hao},
  booktitle={Proceedings of the IEEE/CVF conference on computer vision and pattern recognition},
  pages={21736--21746},
  year={2023}
}

@article{zhou2023partslip++,
  title={Partslip++: Enhancing low-shot 3d part segmentation via multi-view instance segmentation and maximum likelihood estimation},
  author={Zhou, Yuchen and Gu, Jiayuan and Li, Xuanlin and Liu, Minghua and Fang, Yunhao and Su, Hao},
  journal={arXiv preprint arXiv:2312.03015},
  year={2023}
}

@article{chen2025autopartgen,
  title={AutoPartGen: Autogressive 3D Part Generation and Discovery},
  author={Minghao Chen and Jianyuan Wang and Roman Shapovalov and Tom Monnier and Hyunyoung Jung and Dilin Wang and Rakesh Ranjan and Iro Laina and Andrea Vedaldi},
  journal={arXiv preprint arXiv:2507.13346},
  year={2025}
}

@inproceedings{chen2025partgen,
  title={Partgen: Part-level 3d generation and reconstruction with multi-view diffusion models},
  author={Chen, Minghao and Shapovalov, Roman and Laina, Iro and Monnier, Tom and Wang, Jianyuan and Novotny, David and Vedaldi, Andrea},
  booktitle={Proceedings of the Computer Vision and Pattern Recognition Conference},
  pages={5881--5892},
  year={2025}
}

@inproceedings{geng2023gapartnet,
  title={Gapartnet: Cross-category domain-generalizable object perception and manipulation via generalizable and actionable parts},
  author={Geng, Haoran and Xu, Helin and Zhao, Chengyang and Xu, Chao and Yi, Li and Huang, Siyuan and Wang, He},
  booktitle={Proceedings of the IEEE/CVF Conference on Computer Vision and Pattern Recognition},
  pages={7081--7091},
  year={2023}
}

@article{tang2024partpacker,
  title={Efficient Part-level 3D Object Generation via Dual Volume Packing},
  author={Tang, Jiaxiang and Lu, Ruijie and Li, Zhaoshuo and Hao, Zekun and Li, Xuan and Wei, Fangyin and Song, Shuran and Zeng, Gang and Liu, Ming-Yu and Lin, Tsung-Yi},
  journal={arXiv preprint arXiv:2506.09980},
  year={2025}
}

@misc{ma2025p3sam,
	title={P3-SAM: Native 3D Part Segmentation}, 
	author={Changfeng Ma and Yang Li and Xinhao Yan and Jiachen Xu and Yunhan Yang and Chunshi Wang and Zibo Zhao and Yanwen Guo and Zhuo Chen and Chunchao Guo},
	year={2025},
	eprint={2509.06784},
	archivePrefix={arXiv},
	primaryClass={cs.CV},
	url={https://arxiv.org/abs/2509.06784}, 
}

@misc{oquab2023dinov2,
  title={DINOv2: Learning Robust Visual Features without Supervision},
  author={Oquab, Maxime and Darcet, Timothée and Moutakanni, Theo and Vo, Huy V. and Szafraniec, Marc and Khalidov, Vasil and Fernandez, Pierre and Haziza, Daniel and Massa, Francisco and El-Nouby, Alaaeldin and Howes, Russell and Huang, Po-Yao and Xu, Hu and Sharma, Vasu and Li, Shang-Wen and Galuba, Wojciech and Rabbat, Mike and Assran, Mido and Ballas, Nicolas and Synnaeve, Gabriel and Misra, Ishan and Jegou, Herve and Mairal, Julien and Labatut, Patrick and Joulin, Armand and Bojanowski, Piotr},
  journal={arXiv:2304.07193},
  year={2023}
}

@article{connell1987generating,
  title={Generating and generalizing models of visual objects},
  author={Connell, Jonathan H and Brady, Michael},
  journal={Artificial Intelligence},
  volume={31},
  number={2},
  pages={159--183},
  year={1987},
  publisher={Elsevier}
}

@article{10.1007/s00371-007-0197-5,
author = {Shapira, Lior and Shamir, Ariel and Cohen-Or, Daniel},
title = {Consistent mesh partitioning and skeletonisation using the shape diameter function},
year = {2008},
issue_date = {March 2008},
publisher = {Springer-Verlag},
address = {Berlin, Heidelberg},
volume = {24},
number = {4},
issn = {0178-2789},
url = {https://doi.org/10.1007/s00371-007-0197-5},
doi = {10.1007/s00371-007-0197-5},
journal = {Vis. Comput.},
month = mar,
pages = {249–259},
numpages = {11}
}

@article{jones2020shapeAssembly,
	title={ShapeAssembly: Learning to Generate Programs for 3D Shape Structure Synthesis},
	author={Jones, R. Kenny and Barton, Theresa and Xu, Xianghao and Wang, Kai and Jiang, Ellen and Guerrero, Paul and Mitra, Niloy and Ritchie, Daniel},
	journal={ACM Transactions on Graphics (TOG), Siggraph Asia 2020},
	volume={39},
	number={6},
	pages={Article 234},
 	year={2020},
	publisher={ACM}
    }

@article{10.1145/1409060.1409098,
author = {Golovinskiy, Aleksey and Funkhouser, Thomas},
title = {Randomized cuts for 3D mesh analysis},
year = {2008},
issue_date = {December 2008},
publisher = {Association for Computing Machinery},
address = {New York, NY, USA},
volume = {27},
number = {5},
issn = {0730-0301},
url = {https://doi.org/10.1145/1409060.1409098},
doi = {10.1145/1409060.1409098},
journal = {ACM Trans. Graph.},
month = dec,
articleno = {145},
numpages = {12}
}

@article{hanocka2019meshcnn,
  title={MeshCNN: A Network with an Edge},
  author={Hanocka, Rana and Hertz, Amir and Fish, Noa and Giryes, Raja and Fleishman, Shachar and Cohen-Or, Daniel},
  journal={ACM Transactions on Graphics (TOG)},
  volume={38},
  number={4},
  pages = {90:1--90:12},
  year={2019},
  publisher={ACM}
}

@inProceedings{abstractionTulsiani17,
  title={Learning Shape Abstractions by Assembling Volumetric Primitives},
  author = {Shubham Tulsiani and Hao Su and Leonidas J. Guibas and Alexei A. Efros and Jitendra Malik},
  booktitle={Computer Vision and Pattern Regognition (CVPR)},
  year={2017}
}

@misc{lee2025perceptualtaxonomyevaluatingguiding,
      title={Perceptual Taxonomy: Evaluating and Guiding Hierarchical Scene Reasoning in Vision-Language Models}, 
      author={Jonathan Lee and Xingrui Wang and Jiawei Peng and Luoxin Ye and Zehan Zheng and Tiezheng Zhang and Tao Wang and Wufei Ma and Siyi Chen and Yu-Cheng Chou and Prakhar Kaushik and Alan Yuille},
      year={2025},
      eprint={2511.19526},
      archivePrefix={arXiv},
      primaryClass={cs.CV},
      url={https://arxiv.org/abs/2511.19526}, 
}

@article{
paul2025gaussian,
title={Gaussian Scenes: Pose-Free Sparse-View Scene Reconstruction using Depth-Enhanced Diffusion Priors},
author={Soumava Paul and Prakhar Kaushik and Alan Yuille},
journal={Transactions on Machine Learning Research},
issn={2835-8856},
year={2025},
url={https://openreview.net/forum?id=yp1CYo6R0r},
note={}
}

@inproceedings{
kaushik2024sourcefree,
title={Source-Free and Image-Only Unsupervised Domain Adaptation for Category Level Object Pose Estimation},
author={Prakhar Kaushik and Aayush Mishra and Adam Kortylewski and Alan Yuille},
booktitle={The Twelfth International Conference on Learning Representations},
year={2024},
url={https://openreview.net/forum?id=UPvufoBAIs}
}

@article{tversky84,
author = {Tversky, Barbara and Hemenway, Kathleen},
year = {1984},
month = {06},
pages = {169-193},
title = {Objects, parts, and categories},
volume = {113},
journal = {Journal of Experimental Psychology: General},
doi = {10.1037//0096-3445.113.2.169}
}

@article{Stelzner2021Decomposing3S,
  title={Decomposing 3D Scenes into Objects via Unsupervised Volume Segmentation},
  author={Karl Stelzner and Kristian Kersting and Adam R. Kosiorek},
  journal={ArXiv},
  year={2021},
  volume={abs/2104.01148},
  url={https://api.semanticscholar.org/CorpusID:233004650}
}

@inproceedings{takmaz2023openmask3d,
  title={{OpenMask3D: Open-Vocabulary 3D Instance Segmentation}},
  author={Takmaz, Ay{\c{c}}a and Fedele, Elisabetta and Sumner, Robert W. and Pollefeys, Marc and Tombari, Federico and Engelmann, Francis},
  booktitle={Advances in Neural Information Processing Systems (NeurIPS)},
year={2023}
}

@article{ha2018world,
  title={World Models},
  author={Ha, David and Schmidhuber, J{\"u}rgen},
  journal={CoRR},
  year={2018}
}

@inproceedings{rt22023arxiv,
    title={RT-2: Vision-Language-Action Models Transfer Web Knowledge to Robotic Control},
    author={Anthony Brohan and Noah Brown and Justice Carbajal and Yevgen Chebotar and Xi Chen and Krzysztof Choromanski and Tianli Ding and Danny Driess and Avinava Dubey and Chelsea Finn and Pete Florence and Chuyuan Fu and Montse Gonzalez Arenas and Keerthana Gopalakrishnan and Kehang Han and Karol Hausman and Alex Herzog and Jasmine Hsu and Brian Ichter and Alex Irpan and Nikhil Joshi and Ryan Julian and Dmitry Kalashnikov and Yuheng Kuang and Isabel Leal  and Lisa Lee and Tsang-Wei Edward Lee and Sergey Levine and Yao Lu and Henryk Michalewski and Igor Mordatch and Karl Pertsch and Kanishka Rao and Krista Reymann and Michael Ryoo and Grecia Salazar and Pannag Sanketi and Pierre Sermanet and Jaspiar Singh and Anikait Singh and Radu Soricut and Huong Tran and Vincent Vanhoucke and Quan Vuong and Ayzaan Wahid and Stefan Welker and Paul Wohlhart and  Jialin Wu and Fei Xia and Ted Xiao and Peng Xu and Sichun Xu and Tianhe Yu and Brianna Zitkovich},
    booktitle={arXiv preprint arXiv:2307.15818},
    year={2023}
}
}
\appendix
\clearpage

\twocolumn[
\begin{center}
{\Large\bfseries Supplementary}
\end{center}
\vspace{1em}
]

\section{Architecture}
\label{sec+apx:arch}

\subsection{Dense Feature Fusion Module}
\label{sec_apx:featfusion}
The raw geometric features $\mathbf{f}_i^g$ and appearance features $\mathbf{f}_i^a$ capture complementary information: geometry encodes shape characteristics, while appearance provides texture and visual cues. We fuse these modalities through a bi-directional cross-attention module that operates on a $k$-nearest neighbor (KNN) graph to maintain computational tractability. 

\noindent
Given the KNN graph structure with indices $\mathcal{N}_i$ denoting the $k$-nearest neighbors of point $i$, we compute: \\

\noindent
\textbf{Relative Positional Bias.}
To incorporate 3D spatial structure through Fourier-encoded relative positional biases, for each neighbor pair $(i, j)$ where $j \in \mathcal{N}_i$, we compute:
\begin{align}
\mathbf{d}_{ij} &= \mathbf{x}_j - \mathbf{x}_i \in \mathbb{R}^3 \\
\mathbf{f}_{ij} &= \mathbf{d}_{ij} \odot \omega \in \mathbb{R}^{3 \times F} \\
\mathbf{h}_{ij} &= [\sin(\mathbf{f}_{ij}), \cos(\mathbf{f}_{ij})] \in \mathbb{R}^{3 \times 2F} \\
\mathbf{b}_{ij} &= \text{MLP}([\mathbf{d}_{ij}, \text{flatten}(\mathbf{h}_{ij})]) \in \mathbb{R}^H
\end{align}
where $\omega = [2^0, 2^1, \ldots, 2^{F-1}]$ with F=6 are logarithmically-spaced frequencies, and the MLP consists of two layers: $\mathbb{R}^{39} \rightarrow \mathbb{R}^{64} \rightarrow \mathbb{R}^H$ with ReLU activation.

\paragraph{Bi-Directional Cross-Attention.}
Let $H=8$ be the number of attention heads and $d_h = d_m / H = 96$ be the head dimension where $d_m=768$ is the model dimension.

\paragraph{Geometric-to-Appearance Direction:} For each point $i$ with neighbors $\mathcal{N}_i$, geometric features attend to appearance features of neighbors, producing cross-modal features that capture appearance information. 
\begin{align}
\mathbf{Q}_p^i &= \mathbf{W}_q^p \mathbf{f}_i^g \in \mathbb{R}^{H \times d_h} \\
\mathbf{K}_a^{ij} &= \mathbf{W}_k^a \mathbf{f}_j^a \in \mathbb{R}^{H \times d_h}, \quad \forall j \in \mathcal{N}_i \\
\mathbf{V}_a^{ij} &= \mathbf{W}_v^a \mathbf{f}_j^a \in \mathbb{R}^{H \times d_h}, \quad \forall j \in \mathcal{N}_i \\
\alpha_{pa}^{i,h,j} &= \frac{\exp((\mathbf{Q}_p^i[h] \cdot \mathbf{K}_a^{ij}[h]) / \sqrt{d_h} + \mathbf{b}_{ij}[h])}{\sum_{j' \in \mathcal{N}_i} \exp((\mathbf{Q}_p^i[h] \cdot \mathbf{K}_a^{ij'}[h]) / \sqrt{d_h} + \mathbf{b}_{ij'}[h])} \\
\mathbf{z}_p^{i,h} &= \sum_{j \in \mathcal{N}_i} \alpha_{pa}^{i,h,j} \mathbf{V}_a^{ij}[h] \\
\mathbf{r}_p^i &= \mathbf{W}_{pa} \text{concat}_h[\mathbf{z}_p^{i,h}] \in \mathbb{R}^{d_g}
\end{align}

\paragraph{Appearance-to-Geometric Direction:} Symmetrically, appearance features attend to geometric features of neighbors, capturing geometric information. This produces $\mathbf{r}_a^i \in \mathbb{R}^{d_a}$.

\paragraph{Gated Fusion.} Learned sigmoid gates control how much of this cross-modal information to incorporate into each original feature, based on both the original feature and the attended information.
\begin{align}
\mathbf{g}_p^i &= \sigma(\mathbf{W}_g^p [\mathbf{f}_i^g; \mathbf{r}_p^i]) \in \mathbb{R}^{d_g} \\
\mathbf{g}_a^i &= \sigma(\mathbf{W}_g^a [\mathbf{f}_i^a; \mathbf{r}_a^i]) \in \mathbb{R}^{d_a} \\
\tilde{\mathbf{f}}_i^g &= \text{LayerNorm}(\mathbf{f}_i^g + \mathbf{g}_p^i \odot \mathbf{r}_p^i) \\
\tilde{\mathbf{f}}_i^a &= \text{LayerNorm}(\mathbf{f}_i^a + \mathbf{g}_a^i \odot \mathbf{r}_a^i)
\end{align}

\paragraph{Final Projection.} After gated addition and layer normalization, we concatenate both modalities and project through a two-layer MLP to produce fused features $\mathbf{h}_i \in \mathbb{R}^{1216}$ for each point.
\begin{equation}
\mathbf{h}_i = \mathbf{W}_2 \text{GELU}(\mathbf{W}_1 \text{LayerNorm}([\tilde{\mathbf{f}}_i^g; \tilde{\mathbf{f}}_i^a]))
\end{equation}
where $\mathbf{W}_1 \in \mathbb{R}^{d_f \times (d_g + d_a)}$ and $\mathbf{W}_2 \in \mathbb{R}^{d_f \times d_f}$ with $d_g=448$, $d_a=768$, and $d_f=256$. BiCo Fusion operates on local $k{=}16$ nearest neighbor graphs in 3D coordinate space, reducing complexity to $\mathcal{O}(Nk)$.

\begin{figure*}[!htbp]
    \centering
    \caption{\textbf{Part Retrieval Comparison with Find3D.} We demonstrate text-driven part retrieval on two representative objects (airplane and motorbike) from Objaverse-General. Given natural language part queries (e.g., ``body'', ``wing'', ``gas tank'', ``wheel''), ALIGN-Parts identifies and retrieves spatially coherent point groups corresponding to each part. Compared to Find3D (left), our method produces more semantically and spatially consistent part retrievals by leveraging the hierarchical point $\rightarrow$ partlet $\rightarrow$ part label decomposition. This design encourages the discovery of well-connected, semantically meaningful regions rather than fragmented point clusters. Ground truth part segmentations (right) show the target labels. ALIGN-Parts achieves results that closely align with ground truth, validating the effectiveness of our partlet-based formulation for open-vocabulary part localization and retrieval.}
    \includegraphics[width=.8\linewidth]{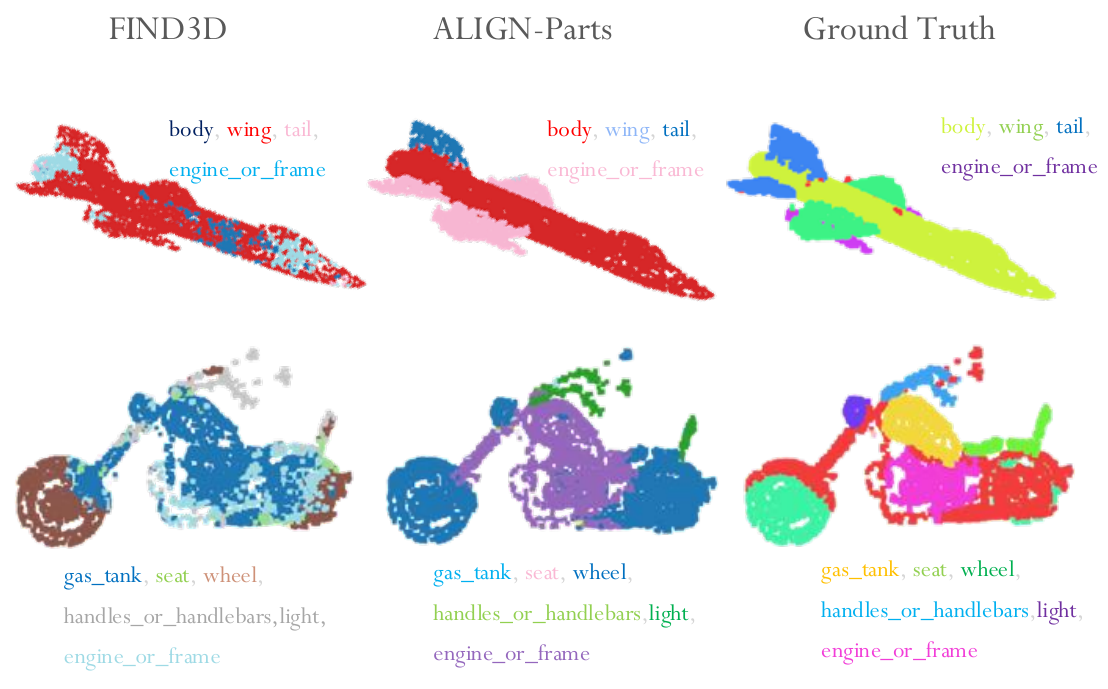}
    \label{fig:part_retrieval}
\end{figure*}

\section{Experiments and Analysis}

Given the challenges inherent in semantic 3D part segmentation, we find that no current published work is directly comparable to our method. To enable rigorous evaluation, we introduce our own strong baseline detailed in \cref{sec:bridge}. While we do include comparisons against class-agnostic 3D part segmentation methods in this manuscript, it is important to note that these do not constitute an entirely fair benchmark for our approach. Most prior methods have been trained using proprietary, closed-source Objaverse-scale datasets, with specific data details and part annotations rarely disclosed publicly. 

In contrast, our experiments are conducted on fully open, publicly available datasets, and our methodology itself improves upon these resources, making our results more easily reproducible and comparable for future researchers. Furthermore, a key emphasis of our approach is efficiency: we process only 10,000 input points per shape, in stark contrast to the 100,000 points typically used by class-agnostic segmentation baselines. This restriction stems from the academic compute limitations we faced, while prior works often benefit from corporate-scale GPU resources.

Despite these constraints, our method achieves competitive or superior performance relative to existing baselines. It is reasonable to expect that, if provided with similar data volumes and computational resources, ALIGN-Parts would further extend its advantage on standard metrics and benchmarks. Our design choices thus not only democratize research in 3D part segmentation but also highlight the promise of reproducibility, accessibility, and efficiency for large-scale semantic understanding in open 3D datasets.

\subsection{PartField+MPNet baseline}
\label{sec:bridge}
Given that our task of semantic part segmentation (in contrast to the relatively easier and more prominent class agnostic part segmentation, we create our own baseline - PartField + MPNet, which assigns labels to parts obtained by KMeans clustering on per-point features. We experimented with two variants of this model, in terms of input features: PartField and PartField + DINOV2, and found that the latter usually yields much better performance. So, without loss of generality, our baseline PartField + MPNet refers to the model where we have per-part PartField + DINOv2 features fused through cross-attention. Specifically, we employ cross-attention fusion to combine per-part geometric (448-D) and appearance (768-D) features, projecting them through 512-D hidden layers into a shared 256-D latent space. The architecture consists of a dense feature fuser (2.8M parameters) with 4 attention heads operating at 512-D, followed by dedicated MLP projectors for local part features (0.39M), semantic text embeddings (0.52M), and global shape descriptors (0.75M), totaling approximately 5.1M parameters. Training optimizes three objectives: symmetric InfoNCE loss for local part-text alignment, a global-level contrastive loss between shape and class embeddings, and a cross-entropy clustering loss that predicts part counts with equal weighting ($\lambda$=1.0) across all terms. The model is trained for 100 epochs using AdamW with learning rate 3e-4, weight decay 1e-5, and cosine annealing schedule ($\eta_\text{min}$=5e-6) with batch size 64. The part count prediction head (0.63M parameters) uses a two-layer MLP with GELU activation to classify the number of semantic parts from fused global features. All projectors and attention mechanisms utilize dropout regularization (p = 0.1) to prevent overfitting during training. During inference, PartField + MPNet first predicts object category by comparing the projected global feature against all class embeddings, then performs soft k-means clustering ($k$ from the part count head) on fused point-level features with Hungarian matching to assign semantic labels by computing cosine similarity between projected cluster centroids and MPNet embeddings of candidate part names.
\begin{figure*}[!htbp]
    \centering
\includegraphics[width=\linewidth]{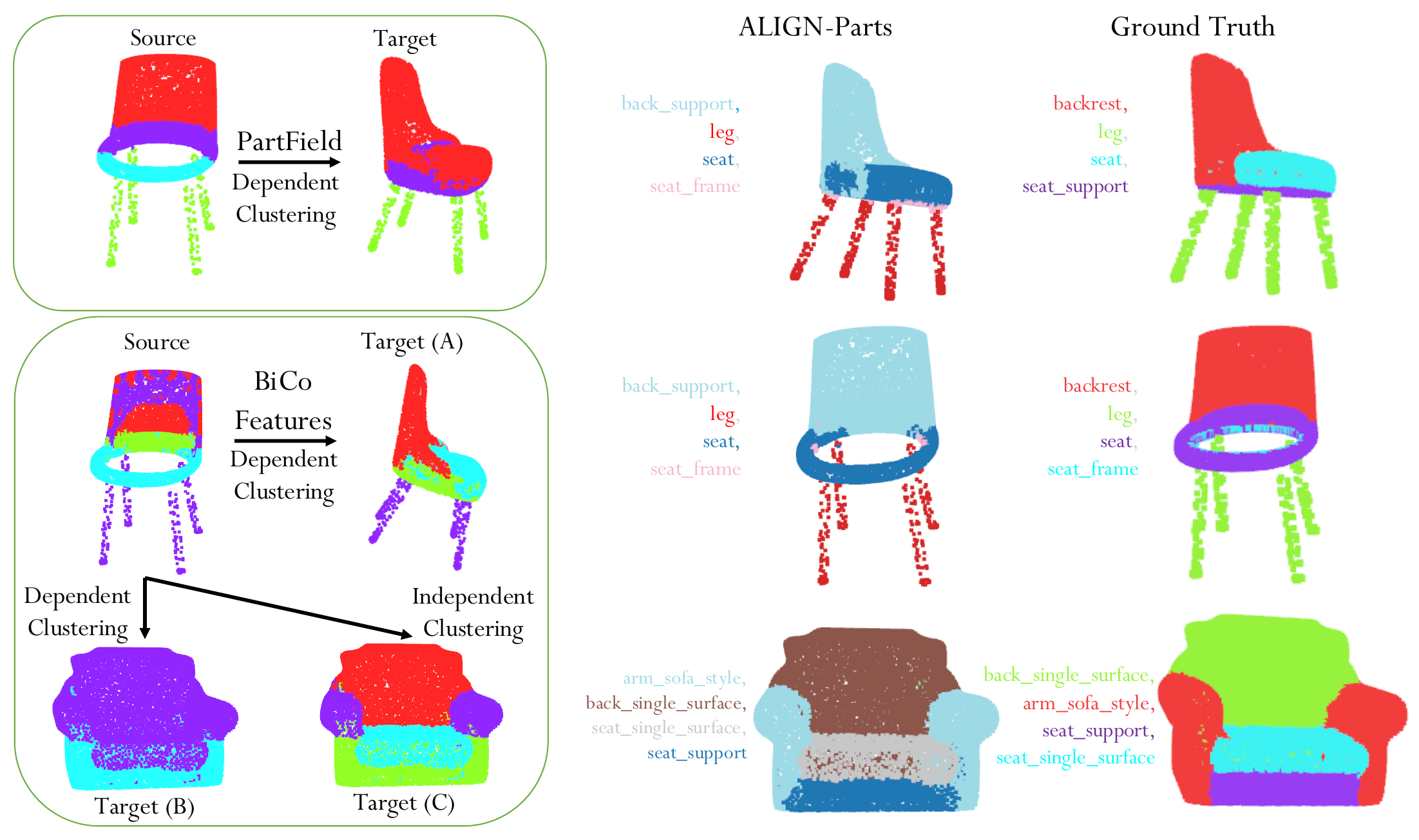}
\caption{\textbf{3D Shape Co-Segmentation Analysis.} \textit{Left: clustering-based co-segmentation.} Prior methods such as PartField perform \emph{dependent clustering} by first segmenting a source shape via feature clustering and then using the resulting cluster means to initialize K-means on a target shape, implicitly enforcing part correspondence; this can break when the target has a different part count or geometry, causing errors such as the red backrest region bleeding into the seat on the target chair. Using the same dependent co-segmentation strategy with our fused BiCo features (\textbf{BiCo Feature Dependent Clustering}) yields improved transfers on moderately similar targets (Target A), but performance degrades on more challenging targets with greater variation in part structure (Target B). As an alternative, we apply \textbf{independent clustering} to Target C, where the target is segmented with source initialization, and clusters are matched post hoc by comparing source and target cluster centers, which proves more reliable for difficult co-segmentation cases. \textit{Middle and right: feedforward ALIGN-Parts.} In contrast to all clustering-based variants, the proposed feedforward ALIGN-Parts model (middle) directly predicts part segmentation and names, achieving robust results across shapes with differing part counts and topologies, and eliminating any dependence on source shapes or explicit co-segmentation.}
\label{fig:coseg}
\end{figure*}

\subsection{Part-Retrieval Comparison with Find3D}

Beyond semantic segmentation, ALIGN-Parts also supports text-driven part retrieval—the task of localizing and retrieving point cloud regions corresponding to natural language part queries. This capability, introduced by Find3D, enables flexible, open-vocabulary part discovery directly from unstructured text descriptions. Our approach performs retrieval by constraining the candidate label vocabulary to only those parts known to be present in the target object class, rather than the full semantic vocabulary. Additionally, we set the number of active partlets to match the ground-truth part count for the object, which serves as an oracle constraint. While this restriction reduces the search space and assignment ambiguity, allowing the model to match predicted part slots to a small, object-specific set of valid labels rather than choosing from dozens of candidates, it also enables fairer and more interpretable comparisons.

This constrained retrieval setup typically yields higher segmentation accuracy by minimizing false positive label assignments and focusing the model's attention on semantically coherent parts. The key advantage of our approach lies in the compositional three-level hierarchy: point cloud $\rightarrow$ partlet $\rightarrow$ part label. This formulation naturally encourages the discovery of connected point groups with consistent semantic meaning, whereas alternatives may suffer from fragmentation or over-segmentation.

We present qualitative comparisons with Find3D on two representative 3D objects from the \emph{airplane} and \emph{motorbike} object classes in the Objaverse-General benchmark~\cite{ma2025find} (part of our closed-vocabulary evaluation set). As shown in \cref{fig:part_retrieval}, ALIGN-Parts consistently retrieves more spatially coherent and semantically meaningful part groups, demonstrating the effectiveness of our partlet-based design for part localization and retrieval tasks.

\subsection{3D Shape Co-Segmentation and Part Label Transfer}
\label{subsec:label_transfer}

\cref{fig:coseg} shows results and analysis of 3D Shape Co-Segmentation using ALIGN-Parts (and the BiCo features) as compared to PartField. A classical approach to 3D part segmentation operates in a co-segmentation setting, where multiple shapes from the same category are jointly analyzed to establish consistent part correspondence. Prior methods, including PartField, employ what we call \emph{dependent clustering}: they first segment a source shape using feature clustering, then initialize k-means clustering on a target shape using the source cluster centroids, implicitly enforcing part correspondence. While this strategy can succeed on geometrically similar shapes, it proves fragile when target shapes exhibit different part counts or topologies. For example, in \cref{fig:coseg}, dependent clustering on a moderately similar target chair (Target A) produces reasonable results, but fails dramatically on targets with substantial part variation (Target B), causing geometric boundaries to blur (e.g., the backrest merging incorrectly with the seat).

An alternative, \emph{independent clustering} approach segments each target shape autonomously and then matches clusters post hoc by comparing source and target cluster centers. As shown in Target C, this mode is more robust to topological differences, though it forgoes any direct geometric correspondence to the source.

In contrast to both clustering-based paradigms, our proposed ALIGN-Parts adopts a fully \emph{feedforward}, discriminative approach that predicts part segmentation masks and semantic labels jointly, without requiring source shape initialization or explicit co-segmentation. This design eliminates brittleness to part count variation and geometric mismatches, enabling robust generalization across shapes with diverse part structures and semantics. As demonstrated in \cref{fig:coseg} (middle and right panels), ALIGN-Parts consistently produces accurate, semantically grounded part segmentations regardless of target shape complexity.

\section{Applications}
Named 3D part decompositions enable a range of downstream applications beyond segmentation benchmarks. ALIGN-Parts can serve as a scalable annotation and taxonomy-normalization engine for dataset construction, complementing part-annotated resources such as PartNet, 3DCoMPaT/3DCoMPaT++, and GAPartNet, which rely on inconsistent or category-specific taxonomies. In robotics and embodied AI, semantic part names aligned with affordances (e.g., handle, support, hinge) provide a natural interface for manipulation and task planning, and can act as structured perceptual inputs to vision-language-action (VLA) models that require interpretable, compositional representations~\cite{rt22023arxiv}. More broadly, named parts offer a compact, semantically grounded state abstraction for world models, where objects and scenes are represented as compositions of interacting components rather than monolithic entities ~\cite{ha2018world}. This representation is particularly relevant for scene understanding and visual question answering (VQA), where prior work on perceptual taxonomies and compositional reasoning argues that structured intermediate representations improve robustness, generalization, and interpretability~\cite{lee2025perceptualtaxonomyevaluatingguiding}. In 3D reconstruction~\cite{paul2025gaussian} and understanding, part-aware representations can support structured reconstruction, correspondence, and pose estimation~\cite{kaushik2024sourcefree} by enforcing semantic consistency across views and instances. Finally, in 3D editing and content creation, named parts enable targeted edits and modular control (e.g., modifying or replacing specific components), complementing recent part-based generative models that discover structure but lack open-vocabulary semantic naming~\cite{chen2025autopartgen}. By producing complete, permutation-consistent named part decompositions in a single feed-forward pass, ALIGN-Parts provides a general-purpose semantic substrate for scene reasoning, VQA, embodied decision-making, and 3D content manipulation without requiring prompts, predefined vocabularies, or task-specific pipelines.

\end{document}